\documentclass{new_tlp}
\usepackage{graphicx}
\usepackage{mathptmx}
\usepackage[most]{tcolorbox}
\usepackage{makeidx}
\usepackage[english]{babel}
\usepackage{amsmath}
\usepackage{amsfonts}
\usepackage{amssymb}
\usepackage{graphicx}
\usepackage{xspace}
\usepackage{url}
\usepackage{algorithm}
\usepackage{algpseudocode}
\usepackage{booktabs}
\usepackage{multirow}
\usepackage[labelsep=quad,indention=10pt]{subfig}
\usepackage{bbding}
\usepackage{todonotes}
\usepackage{adjustbox}
\usepackage{rotating}
\usepackage{paralist}
\usepackage{float}

\newcommand{\sysfont}{\textit}

\newcommand{\dlv}{\sysfont{DLV}\xspace}
\newcommand{\idlv}{{{\small $\cal I$}-}\dlv}
\newcommand{\iidlv}{{{\small $\cal I$}$^2$-}\dlv}

\newcommand{\system}{\sysfont{I-DLV-sr}\xspace}

\newcommand{\flink}{\sysfont{Flink}\xspace}

\newcommand{\derives}{\mbox{\,:\hspace{0.1em}\texttt{-}}\,\xspace}

\newcommand{\cit}[1]{~\cite{#1}}
\newcommand{\re}[1]{~\ref{#1}}

\newcommand{\commentsymbolright}{/*}
\newcommand{\commentsymbolleft}{*/}

\algrenewcommand\algorithmiccomment[1]{\hfill \commentsymbolright{} #1 \commentsymbolleft{}}


\newtheorem{example}{Example}
\newcommand{\finex}{{\footnotesize\hfill \ensuremath{\Box}}}
\newenvironment{monospace}{\fontfamily{qcr}\selectfont}{\par}
\newenvironment{larscode}[1]{
\begin{tcolorbox}[top=0pt, bottom=0pt, left=2pt,right=2pt,enhanced,opacityframe=.05]
\begin{footnotesize}
\begin{monospace}}{\end{monospace}
\end{footnotesize}\end{tcolorbox}}

\newcommand{\quot}[1]{``#1"}

\makeatletter
\let\@fnsymbol\@arabic
\makeatother

\algdef{SE}[DOWHILE]{Do}{doWhile}{\algorithmicdo}[1]{\algorithmicwhile\ #1}%

\newcommand{\EM}{\textit{Execution Manager}\xspace}
\newcommand{\SM}{\textit{Stream Manager}\xspace}
\newcommand{\CM}{\textit{Subprogram Manager}\xspace}

\newcommand{\mono}[1]{\begin{footnotesize}{\fontfamily{qcr}\selectfont#1}\end{footnotesize}}

\newcommand{\mapping}{$\tau$\xspace}

\newtheorem{prop}{Proposition}
\newcommand{\sa}{\emph{streaming atoms}\xspace}

\newcommand{\atleast}[2]{\textbf{at least} \textit{#1} \textbf{in [}\textit{#2}\textbf{]}}
\newcommand{\countop}[2]{\textbf{count} \textit{#1} \textbf{in [}\textit{#2}\textbf{]}}
\newcommand{\countopd}[2]{\textbf{count} \textit{#1} \textbf{in \{}\textit{#2}\textbf{\}}}
\newcommand{\atleastd}[2]{\textbf{at least} \textit{#1} \textbf{in \{}\textit{#2}\textbf{\}}}
\newcommand{\atmost}[2]{\textbf{at most} \textit{#1} \textbf{in [}\textit{#2}\textbf{]}}
\newcommand{\always}[1]{\textbf{always in [}\textit{#1}\textbf{]}}
\newcommand{\fin}[1]{\textbf{in [}\textit{#1}\textbf{]}}
\newcommand{\mgp}{$P$\xspace}
\newcommand{\gp}{P\xspace}
\newcommand{\msdg}{$G_{\gp}^{SD}$\xspace}
\newcommand{\sdg}{G_{\gp}^{SD}\xspace}
\newcommand{\mscg}{$G_{\gp}^{SC}$\xspace}
\newcommand{\scg}{G_{\gp}^{SC}\xspace}
\newcommand{\mmod}[1]{M^{#1}\xspace}
\newcommand{\mmmod}[1]{$\mmod{#1}$\xspace}
\newcommand{\al}{\;\mathbf{at \;least}\;}
\newcommand{\am}{\;\mathbf{at \;most}\;}
\newcommand{\alwm}{\;\mathbf{always}\;}
\newcommand{\inm}{\;\mathbf{in}\;}
\def\mmedia{\textit{Content Caching}\xspace}
\def\pvsystem{\textit{Photo-voltaic System}\xspace}
\def\hjoin{\textit{Heavy Join}\xspace}
\def\time{\texttt{total time}\xspace}

\def\req{\texttt{\#accepted requests}\xspace}
\def\lat{\texttt{latency}\xspace}

\usepackage[normalem]{ulem}
\newif\ifrevisionmode
\revisionmodefalse

\ifrevisionmode
    \newcommand{\add}[1]{{\color{blue}#1}}
    \newcommand{\remove}[1] { {\color{red}\sout{#1}} }
    \newcommand{\replace}[2]{{\color{red}\sout{#1}}{\color{blue}#2}}
\else
    \newcommand{\add}[1]{#1}
    \newcommand{\remove}[1]{}
    \newcommand{\replace}[2]{#2}
\fi

\title[I-DLV-sr: A Stream Reasoning System based on I-DLV]{I-DLV-sr: A Stream Reasoning System based on I-DLV\thanks{This work has been partially supported by the project ``MAP4ID - Multipurpose Analytics Platform 4 Industrial Data'', N. F/190138/01-03/X44 and by the Italian MIUR Ministry and the Presidency of the
Council of Ministers under the project \quot{Declarative Reasoning over
Streams} under the \quot{PRIN} 2017 call (CUP $H24I17000080001$, project
2017M9C25L\_001).}}

  \author[F. Calimeri et al.]
         {
         FRANCESCO CALIMERI$^{1}$, MARCO MANNA$^{1}$, ELENA MASTRIA$^{1}$
         \and  MARIA CONCETTA MORELLI$^{2}$
         , SIMONA PERRI$^{1}$
         , JESSICA ZANGARI$^{1}$\\
          Department of Mathematics and Computer Science, University of Calabria, Rende, Italy\\
         \email{$^{1}\{$name.surname$\}$@unical.it, $^{2}$maria.morelli@unical.it}
        }
\jdate{}
\pubyear{}
\submitted{18 May 2021}
\revised{14 July 2021}
\accepted{31 July 2021}

\begin{document}

\label{firstpage}

\maketitle

  \begin{abstract}
    We introduce a novel logic-based system for reasoning over data streams, which relies on a framework enabling a tight, fine-tuned interaction between \textit{Apache Flink} and the \iidlv system.
    The architecture allows to take advantage from both the powerful distributed stream processing capabilities of \flink and the incremental reasoning capabilities of \iidlv, based on overgrounding techniques.
    Besides the system architecture, we illustrate the supported input language and its modeling capabilities, and discuss the results of an experimental activity aimed at assessing the viability of the approach. This paper is under consideration in Theory and Practice of Logic Programming (TPLP).
   \end{abstract}

  \begin{keywords}
    Stream Reasoning, Stream Processing, 
    Overgrounding, Knowledge Representation and Reasoning, Answer Set Programming
  \end{keywords}

%
%
\section{Introduction}\label{sec:intro}

		



Stream Reasoning (SR)\cit{DBLP:journals/datasci/DellAglioVHB17} consists in the application of inference techniques to data streams.
Recently, SR has been studied in several 
fields, and became more and more relevant in diverse application scenarios, such as IoT, Smart Cities, Emergency Management.
Hence, different approaches have been proposed~\cite{DBLP:journals/ijsc/BarbieriBCVG10,DBLP:conf/semweb/PhuocDPH11,hoeksema2011high,DBLP:conf/lpnmr/PhamAM19} in contexts such as Complex Event Processing (CEP), Semantic Web and Knowledge Representation and Reasoning (KRR).
%
Among declarative KRR paradigms, Answer Set Programming (ASP)\cit{DBLP:journals/cacm/BrewkaET11} is a well-established proposal 
that gained attention also outside of the academia thanks to the availability of robust and efficient implementations\cit{DBLP:conf/ijcai/GebserLMPRS18}.
ASP is acknowledged as a particularly attractive basis for SR.
Indeed, some important steps forward in this direction have been taken; we mention here the LARS framework~\cite{beck2018lars} and its implementations~\cite{bazoobandi2017expressive,beck2018ticker,eiter2019distributed,DBLP:conf/bigdataconf/RenCNX18}
StreamRule\cit{DBLP:conf/rr/MileoAPH13}, C-ASP\cit{DBLP:conf/lpnmr/PhamAM19} and others\cit{DBLP:conf/ai/DoLL11,DBLP:conf/lpnmr/GebserGKS11}.
However, ASP-based stream reasoners appear not mature enough with respect to the desirable requirements for SR\cit{DBLP:journals/datasci/DellAglioVHB17}.
Hence, there is still room for improvements, especially when dealing with real applications: for instance, some systems are weakly usable in practice (e.g., enforce strict assumptions in input programs) and others suffer from efficiency/scalability issues.


The aim of our work is to obtain a novel, reliable ASP-based stream reasoner; besides efficiently scaling over real-world application domains, it should support a language which inherits the highly declarative nature and ease of use from ASP, while being also easily extendable with new constructs that are relevant for practical SR scenarios.
In this paper we present the prototype of \system, a system that relies on the proper integration of two well-established solutions in the field of Stream Processing and ASP, respectively.
\system is based on a continuous cooperation between two components: an \replace{ad-hoc}{custom} designed application that leverages on \textit{Apache Flink}~\cite{hueske2019stream}, a powerful stream processor for efficiently managing data streams, and \iidlv~\cite{DBLP:journals/tplp/IanniPZ20}, an ASP grounder and a full-fledged deductive database system that enables incremental ASP evaluation via overgrounding techniques~\cite{CALIMERI_2019}.
Currently, the supported language  
basically consists in normal stratified ASP enriched with a set of constructs allowing to reason over streams.
We tested the system with the aim of assessing its reliability, performance, and scalability, and also of exploring ease of modelling and reasoning capabilities.
The results of a number of experiments\add{,} conducted on both real-world and synthetic domains, are encouraging, proving
the viability of \add{the} approach and \add{the} robustness of  \add{the} implementation.


The remainder of the paper is structured as follows.
In Section~\ref{sec:components} we describe the two main components of \system; Section~\ref{sec:language} defines syntax and semantics of the currently supported language, and illustrates how it can be used for modelling SR problems via proper examples; in Section~\ref{sec:architecture} the architecture of \system is presented and discussed in detail; Section~\ref{sec:experiments} illustrates the experimental settings and the results; related works are discussed in Section~\ref{sec:related_works}, and, eventually, conclusions and future works are reported in Section~\ref{sec:future_work}.

\section{\system Components}\label{sec:components}
\system mainly consists of two components: a \replace{ad-hoc designed \flink}{custom \flink-based Java} application and \iidlv.
\flink is a distributed Stream Processing system for both batch and real-time stream data processing with high throughput and low latency.
Applications relying on \flink are built by designing, using the exposed APIs, ad-hoc dataflow graphs that make use of a number of different operators; besides pre-defined ones, custom operators can also be implemented.
Each operator transforms one or more input data streams into a new data stream, that, in turn, can be the part of the input to subsequent operators.
%
A dataflow graph is basically a directed acyclic graph, where nodes represent the operators, incoming arcs represent the input data streams, and outcoming arcs represent the intermediate data streams resulting from operator applications.
Each dataflow graph must have one or more sources, from which the data streams originate, and at least one sink, 
that can either emit the final output or persistently store it.
Computation is automatically distributed and parallelized on the basis of dependencies among operators\cit{carbone2015apache}.

\emph{{$\cal I$}ncremental} \idlv system, namely \iidlv \cit{DBLP:journals/tplp/IanniPZ20},
is the recently presented incremental version of the deductive database system and ASP grounder \idlv\cit{DBLP:conf/aiia/CalimeriFPZ16}.
It incorporates overgrounding techniques\cit{CALIMERI_2019} to accommodate incremental executions
over different inputs.
More in detail, \iidlv works in a server-like mode: given a fixed input program, it remains ``listening'' for input facts.
Every time new such facts arrive, it computes a ground program by properly updating the one resulting from previous ``shots''.
In other words, it automatically maintains a ground program, semantically equivalent to the input one and monotonically growing over time.
It is worth noting that, just like \idlv, in case of normal and stratified w.r.t. negation ASP programs, \iidlv computes the full semantics (i.e., returns the unique answer set).
Typically, an overgrounded program, after a number of shots, converges to a propositional theory general enough to be reused together with possible future inputs, with little or no further update required.
This makes overgrounding very attractive in SR contexts, as grounding activities in later iterations tend to be virtually eliminated.

\section{\system Language}\label{sec:language}
In this section, we introduce the syntax and the semantics of \system programs and then we show the usage of \system language for Knowledge Representation via some example scenarios.


\smallskip
\noindent\textbf{\textit{Syntax.}}\label{par:syntax}
We assume to have finite sets $V$, $C$ and $P$ consisting of \emph{variables}, \emph{constants} and \emph{predicate names}, respectively; we constrain $V$ and $C$ to be disjoint. 
A \emph{term} is either a variable in $V$ or a constant in $C$.
A \emph{predicate atom} has the form $p(t_1,\dots,t_n)$, where $p\in P$ is a predicate name, $t_1,\dots,t_n$ are terms and $n\geq0$ is the arity of the predicate atom; a predicate atom $p()$ of arity $0$ can be also denoted by $p$.
A predicate atom is {\em ground} if none of its terms is a variable.
We denote as $G$ the set of all ground predicate atoms constructible from predicate names in $P$ and constants in $C$.
%
%
%
Given a \emph{predicate atom} $a$, a constant $c\in  C\cap \mathbb{N^+}$, a term $t\in C\cup V$ (\emph{counting term}), and a non-empty set of numbers $D=\{d_1,\dots,d_m\}$ $\subset\mathbb{N}$, we define three types of \sa:
%
%
\\
\indent \indent $ a  \;\mathbf{at \;least}\; c \;\mathbf{in} \; \{ d_1,\dots,d_m\}\ \ \ \ \ \ \ \
a  \;\mathbf{always \; in} \; \{d_1,\dots,d_m\}\ \ \ \ \ \ \ \
a  \;\mathbf{count}\; t \;\mathbf{in} \; \{ d_1,\dots,d_m\}
$\\
A streaming atom $\alpha$ (resp., $\mathtt{not}\ \alpha$) is said to be a \emph{positive streaming literal} (resp., \emph{negative streaming literal}), where $\mathtt{not}$ denotes \emph{negation as failure}.
A streaming literal is said to be {\em ground} if none of its terms is a variable.
For a set $L$ of streaming literals, $preds(L)$ denotes the set of predicates appearing in $L$.
The following shortcuts are admitted:
\begin{compactitem}
\item[-]
    $a \;\mathbf{in} \; \{d_1,\dots,d_m\}$ in place of $a \;\mathbf{at \;least}\; 1 \;\mathbf{in} \; \{d_1,\dots,d_m\} $;
\item[-]
    $a$  \ in place of $a \;\mathbf{at \;least}\; 1 \;\mathbf{in} \; \{0\} $ (this is called ``degenerate'' form of a streaming literal);
\item[-]
    $a \; \mathbf{at \;most}\; c \;\mathbf{in} \; \{d_1,\dots,d_m\}$ in place of $\mathtt{not}\; a  \;\mathbf{at \;least}\; c' \;\mathbf{in} \; \{ d_1,\dots,d_m\} $ where $c'=c+1$.
\end{compactitem}
Given a streaming atom of any type, if $D=\{n \in \mathbb{N}\ \vert\ 0\le n \le w \wedge w>0\}$ we indicate it simply as $[w]$; e.g., we write $a \;\mathbf{always \; in}\;[3] $ instead of $a \;\mathbf{always \; in} \; \{0, 1, 2, 3\}$.

A rule can be of one out of the two forms:
$(1)\ \ a \derives\; l_1,\dots,l_b.$\ \  or $(2)\ \ \mathbf{\#temp} \; a \derives\; l_1,\dots,l_b.$, where $a$ is a predicate atom, $b\ge0$ and $l_1,\dots,l_b$ represent a conjunction of streaming literals.
For a rule $r$, we say that the \emph{head} of $r$ is the set $H(r) = \{a\}$, whereas the set $B(r) = \{l_1,\dots,l_b\}$ is referred to as the \emph{body} of $r$.
A program \mgp is a finite set of rules;
\mgp is \remove{said} \emph{flat} if all rules contain only streaming literal\add{s} in the degenerate form;
\mgp is \remove{said} \emph{restricted} if only rules of form $(1)$ occur in it.
We say that a rule $r$ is \textit{safe} if all variables appearing in $H(r)$ or in a negative streaming literal of $B(r)$ also appear in a positive streaming literal of $B(r)$.
A program is safe if all its rules are safe. We require programs to be safe.


A streaming literal is said to be \textit{harmless} if it has form
$a \;\mathbf{at \;least}\; c \;\mathbf{in} \;  \{d_1,\dots,d_m\} $ or
$a \;\mathbf{always \; in}$ $\;  \{d_1,$ $\dots$,$d_m\}$; otherwise, it is said to be \textit{non-harmless}.
%
A program \mgp is \emph{stratified} if there is a partition of disjoint sets of rules $\gp=\Pi_1\cup \dots\cup\Pi_k$ (called strata) such that for $i\in\{1,\dots,k\}$ both these conditions hold:
     ($i$) for each harmless literal in the body of a rule in $\Pi_i$ with predicate $p$, $\{r \in \gp \vert H(r)=$ $\{p(t_1,\ldots,t_n)\}\}\subseteq \bigcup_{j=1}^{i} \Pi_j$;
    ($ii$) for each non-harmless literal in the body of a rule in $\Pi_i$ with predicate $p$, $\{r \in \gp \vert H(r)=$ $\{p(t_1,\ldots,t_n)\}\}\subseteq \bigcup_{j=1}^{i-1} \Pi_j$.
We call $\Pi_1,\dots,\Pi_k$ a \emph{stratification} for \mgp and \mgp is \remove{said} stratified by $\Pi_1,\dots,\Pi_k$.
An \system program is always stratified.





\smallskip
\noindent\textbf{\textit{Semantics.}}
We \replace{now}{} provide \add{next} an operational semantics of \system programs.
We start by introducing the notion of \add{a} stream.
%
A \emph{stream $\Sigma$} is a sequence of sets of ground predicate atoms $\langle S_0,\dots, S_n\rangle$ such that for $\;0\leq i \leq n$, $S_i\subseteq G$.
Each natural number $i$ is \replace{said}{called} \emph{time point}.
A ground predicate atom $a\in S_i$ is true at \add{the} $i$-th time point.  \add{Given two streams $\Sigma=\langle S_0,\dots, S_n\rangle$ and $\Sigma'=\langle S'_0,\dots, S'_n\rangle$,  $\Sigma=\Sigma'$ iff $S_i=S'_i$ for each $i\in \{0,\dots,n\}$.}
%
For a stream $\Sigma=\langle S_0,\dots, S_n\rangle$, a {\em backward observation} identifies ground predicate atoms that are true at some time points preceding the $n$-th \add{time point}.
More formally, given a stream $\Sigma=\langle S_0,\dots, S_n\rangle$ and a set of numbers $D$\remove{$=\{d_1,\dots,d_m\}$}$\subset \mathbb{N}$, we define the \emph{backward observation of $\Sigma$ w.r.t. $D$} \replace{, denoted \remove{as} $O(\Sigma, D)$,}{as} the set \replace{
$\{S_i\ \vert\ i=n-d_k \text{ with } d_k\in D \wedge i\geq0\}$} {
$\{S_i\ \vert\ i=n-d \text{ with } d\in D \wedge i\geq0\}$, } \add{and we denote it as $O(\Sigma, D)$}.
Given $w\in\mathbb{N}$, a backward observation of $\Sigma$ w.r.t. $[w]$ is called \emph{window}.


A backward observation allows to define the truth of a ground streaming literal at a given time point.
Given a stream $\Sigma=\langle S_0,\dots, S_n\rangle$, $D=\{d_1,\dots,d_m\}\subset \mathbb{N}$, $c\in C \setminus \{0\}$ and the backward observation $O(\Sigma,D)$, Table\re{table:entails} reports when $\Sigma$ \emph{entails} a ground streaming atom $\alpha$ (denoted $\Sigma\models \alpha$) or its negation ($\Sigma\models  \mathtt{not} \; \alpha$). If $\Sigma\models\alpha$ ( $\Sigma\models  \mathtt{not} \; \alpha$) we say that $\alpha$ is true (false) at time point $n$.
\begin{table}[h]
\small
\centering
\begin{tabular}{c c c}
\toprule
$\alpha$ & $\Sigma\models \alpha$ & $\Sigma\models  \mathtt{not} \; \alpha$ \\
\midrule
$a \;\mathbf{at \;least}\; c \;\mathbf{in} \; \{d_1,\dots,d_m\} $ & $\vert\{A\in O(\Sigma,D): a \in A\}\vert\geq c $&$  \vert\{A\in O(\Sigma,D): a \in A\}\vert < c$\\


$a\; \mathbf{always\; in} \;\{d_1,\dots,d_m\}$&$ \forall A\in O(\Sigma,D),  a \in A $ & $\exists A\in O(\Sigma,D): a \not\in A $\\
$a \;\mathbf{count}\; c \;\mathbf{in} \; \{d_1,\dots,d_m\}$ & $\vert\{A\in O(\Sigma,D): a \in A\}\vert = c $&$  \vert\{A\in O(\Sigma,D): a \in A\}\vert \neq c$\\

\bottomrule
\end{tabular}
\caption{Entailment of ground streaming literals.}
\label{table:entails}
\vspace{-0.4cm}
\end{table}
\begin{example}\label{Entailment}
In the stream $\Sigma=\langle\{a(2),b(5)\},\{a(3),c(7)\},\{b(5)\},\{a(3)\}\rangle$, atom $b(5)$ is true at time points $0$ and $2$.
For $D=\{0,1,3\}$, the backward observation of $\Sigma$ w.r.t. $D$ is $O(\Sigma,D)=\{\{a(3)\},$
$\{b(5)\},$ $\{a(2),b(5)\}\}$. We have that $\Sigma\models b(5)\;\mathbf{at \;least}\; 2 \;\mathbf{in} \; \{0,1,3\}$. Indeed, the cardinality of the set $\{A\in O(\Sigma,D): b(5) \in A\}$ is $2$.
\end{example}

We define now the notions of substitution and applicability of a rule to a stream.
%
A \emph{substitution} $\sigma$ is a mapping from the set $V$ of variables to the set of constants $C$. Given a predicate atom $a$ and a \emph{substitution} $\sigma$, $\sigma(a)$ is the ground predicate atom obtained by replacing each occurrence of a variable $v$ in $a$ by $\sigma(v)$. Given a streaming literal $l$, $\sigma(l)$ is the ground streaming literal obtained by applying $\sigma$ to the predicate atom appearing in $l$ and to the counting term $t$ possibly appearing in $l$ if $t\in V$
and $\sigma(t)\ne0$.
Given a stream $\Sigma=\langle S_0,\dots, S_n\rangle$, a rule $r$ is \emph{applicable} on $\Sigma$ if there exists a substitution $\sigma$ such that $\Sigma \models \sigma$ \replace{$(b_i)$}{$(b)$} for all \replace{$b_i$}{$b$}$\in B(r)$. In such a case, $r$ is \remove{said} \emph{applicable} on $\Sigma$ via $\sigma$.
%
Roughly, applicability of rules identifies new ground predicate atoms that are true at time point $n$; that is, a
rule $r$ fires implying the truth of the ground predicate atom $\sigma(a)$, with $a\in H(r)$.
%
%
Given a stream $\Sigma=\langle S_0,\dots, S_n\rangle$ and an \system program \mgp, a \emph{trigger for \mgp on $\Sigma$} is a pair $\langle r, \sigma\rangle$ where $\sigma$ is a substitution and $r\in \gp$ is applicable on $\Sigma$ via $\sigma$.
An \emph{application} of $\langle r,\sigma \rangle$ to $\Sigma$ returns the stream $\Sigma^{\prime}=\langle S_0,\cdots,S_{n-1},S_n\cup\sigma(a)\rangle$, with $a\in H(r)$. A trigger application is denoted as $\Sigma\langle r,\sigma\rangle \Sigma^{\prime}$.

\begin{example}
Consider again $\Sigma$ as in Example\re{Entailment} and let $\mgp_1$ be as follows:
\begin{larscode}

\centering
$r_1$: c(X) \derives b(X) \textbf{at least }\textit{2} \textbf{in \{} \textit{0,1,3}\textbf{ \}}.
\end{larscode}
\noindent Let us consider a substitution $\sigma$ \remove{be} such that $X\mapsto5$.
The rule $r_1$ is applicable on $\Sigma$ via $\sigma$ and the application of $\langle r_1,\sigma\rangle$ to $\Sigma$ returns $\Sigma^{\prime}=\langle\{a(2),b(5)\},\{a(3),c(7)\},\{b(5)\},\{a(3),c(5)\}\rangle$.
\end{example}

We therefore \remove{ define the application of an \system program $\gp$ to a stream as the application of the rules in $\gp$.
We }establish the order of application of the rules \add{of an \system program \mgp} according to a stratification for $P$\add{,} and finally\replace{, we}{} introduce the concept of streaming model of an \system program on a stream.


Given a stream $\Sigma$ and an \system program \mgp \replace{a \emph{program application} of \mgp}{stratified by $\Pi_1 \dots\Pi_k$,
a \emph{stratum application} of $\Pi_s$ for $s\in \{1,\dots,k\}$} on $\Sigma$ is a finite sequence of streams $\Sigma_0, \dots, \Sigma_h$ with $\Sigma_0=\Sigma$ and $h\geq 0$ such that:
\begin{compactitem}
\item[-]
    for each $0 \leq i < h$, there is a trigger $\langle r_i, \sigma_i\rangle$ for \replace{\mgp}{$\Pi_s$} on $\Sigma_i$ such that $\Sigma_i\langle r_i, \sigma_i \rangle \Sigma_{i+1}$;
\item[-]
    for each $0\leq i < j< h$, if $\Sigma_i\langle r_i, \sigma_i \rangle \Sigma_{i+1}$\replace{ and}{,} $\Sigma_j\langle r_j, \sigma_j \rangle \Sigma_{j+1}$\replace{,}{and} $r_i=r_j$, then $\sigma_i \ne \sigma_j$;
\item[-]
    there is no trigger $\langle r, \sigma\rangle$ for \replace{\mgp}{$\Pi_s$} on $\Sigma_h$ such that $\langle r, \sigma\rangle \notin \{\langle r_i, \sigma_i\rangle\}_{0 \leq i \leq h}$.
\end{compactitem}

\add{
Intuitively, starting from a stream $\Sigma$, we apply distinct triggers, considering only different substitutions for the same rule, as long as there is a new trigger.
In general, for each stratum $\Pi_s$ more than one stratum application exists. However, for any two stratum applications of $\Pi_s$ on $\Sigma$, their last streams coincide. This intuition is proved by the following proposition whose proof is given in\re{sec:appendixA}.

\begin{prop}\label{prop:stratum}
Given a stream $\Sigma$, an \system program \mgp stratified by $\Pi_1 \dots\Pi_k$, a stratum $\Pi_s$ with $s\in \{1,\dots,k\}$ and
two stratum applications of $\Pi_s$ on $\Sigma$, $\Sigma_0, \dots, \Sigma_h$ and $\Sigma'_0, \dots, \Sigma'_t$, we have that $\Sigma_h=\Sigma'_t$.
\end{prop}
}
We call the \add{last} stream \replace{$\Sigma_h$}{in a stratum application} \emph{outcome of \replace{the program application of \mgp}{$\Pi_s$} on $\Sigma$}, \replace{and denote it}{denoted} as \replace{$outcome(\gp,\Sigma)$}{$outcome(\Pi_s,\Sigma)$}.
\replace{At this point, for a stratified program \mgp, the outcome over strata of \mgp on a stream $\Sigma$ is defined by considering that the outcome of each stratum depends on the outcomes of program applications of previous strata.}{We now define the outcome over strata of \mgp on $\Sigma$, obtained, starting from the outcome of the first stratum $\Pi_1$ of \mgp and considering, one after the other, subsequent strata according to the stratification.}

Given an \system program \mgp stratified by $\Pi_1 \dots\Pi_k$ and a stream $\Sigma=\langle S_0,\dots, S_n\rangle$, we compute the sequence of streams defined as follows: 
$ \Sigma_{\Pi_1}=
        outcome(\Pi_1, \Sigma) $
        and
    $\Sigma_{\Pi_i}=outcome(\Pi_i, \Sigma_{\Pi_{i-1}})$  \mbox{for } $1 < i\leq k$.
We define \emph{outcome over strata} of an \system program \mgp on the stream $\Sigma$, denoted ${\cal R}(\gp, \Sigma)$, the $n$-th element in the stream $\Sigma_{\Pi_k}$. \add{Note that ${\cal R}(\gp, \Sigma)$ is independent from the chosen stratification.}

\smallskip
\noindent\textit{Streaming model of \system restricted programs.}
For simplicity, we first define the notion of streaming model for programs where only rules of the form $(1)$ appear. \\
%
Let $\Sigma=\langle S_0,\dots, S_n\rangle$ be a stream, \mgp an \system restricted program stratified by $\Pi_1 \dots\Pi_k$. Let $\Sigma^{\prime} = \langle S'_0, \dots, S'_{n-1}, S_n\rangle $ be such that $S'_0=$ ${\cal R}(\gp, \langle S_0\rangle)$ and each $S'_i$ for $i \in \{1,\dots,n-1\}$ is defined as  ${\cal R}(\gp, \langle S'_0,\dots,S'_{i-1},S_i\rangle)$. We define  ${\cal R}(\gp, \Sigma^{\prime})$ as \emph{streaming model of \mgp on $\Sigma$}.

%

Intuitively, the streaming model of \mgp for a stream $\Sigma=\langle S_0,\dots, S_n\rangle$ is the set of the ground predicate atoms derived as true at the $n$-th time point.
These latter come from the evaluation of \mgp over a stream that iteratively accumulates the evaluations at previous time points.

\begin{example}\label{ex:output}
Let $\Sigma=\langle\{b(5)\},\{c(7)\}\rangle$ and $\gp_2$ be the following program:
\begin{larscode}

\centering
$r_1$:\ c(X) \derives b(X). \ \ \ \ \ \ \
$r_2$:\ d(X) \derives c(X) \fin{1}.
\end{larscode}
Note that $\gp_2$ is stratified by the single stratum $\{r_1,r_2\}$; thus, the outcome over strata of $\gp_2$ on a stream $\Sigma$, ${\cal R}(\gp_2, \Sigma)$, is the $n$-th element in the stream $outcome(\gp_2,\Sigma)$.
We have $S'_0=$ ${\cal R}(\gp_2, \langle S_0\rangle)= {\cal R}(\gp_2, \langle\{b(5)\}\rangle)= \{b(5),c(5),d(5)\}$ and $\Sigma^{\prime}=\langle S'_0,S_1\rangle=$  $\langle\{b(5),c(5),d(5)\},\{c(7)\}\rangle$.\ \
Since\ \ ${\cal R}(\gp_2,\Sigma^{\prime})={\cal R}(\gp_2, \langle\{b(5),c(5),d(5)\},\{c(7)\}\rangle)=\{c(7),d(7),d(5)\}$, the streaming model of $\gp_2$ for $\Sigma$ is the set $\{c(7),d(7),d(5)\}$.
\end{example}

\noindent\textit{Streaming model of \system programs.}
We now introduce the streaming model also for \system non-restricted programs.
To this aim, we need 
the notion of persistent outcome and 
a slightly different definition of streaming model that takes into account the eventual presence of $\mathbf{\#temp}$ in rule heads.
Such marking stands for temporary: indeed the truth of heads of rules in the form $(2)$ is in a sense, limited to the current time point.
We are given a stream $\Sigma=\langle S_0,\dots, S_n\rangle$ and an \system program \mgp stratified by $\Pi_1 \dots\Pi_k$.
Let $\gp_{(1)}=\{r \in \gp \vert r \mbox{ is of the form }(1)\}$,
we define \emph{persistent outcome over strata} of \mgp on $\Sigma$ the set of ground predicate atoms $\{a \in$ ${\cal R}(\gp,\Sigma)\ \vert\ a\in S_n \vee\ (\exists r \in \gp_{(1)}  \text{ s.t. } r \mbox{ is applicable to } \Sigma_{\Pi_k} \text{ via}$ $\sigma \land \sigma(h)=a \land h\in H(r))\}$ and we denote it with ${\cal P}(\gp,\Sigma)$.
%
Basically, for an input stream $\Sigma=\langle S_0,\dots, S_n\rangle$, ${\cal P}(\gp,\Sigma)$ identifies the set of ground predicate atoms in ${\cal R}(\gp,\Sigma)$ deriving from heads of rules of the form (1) or in $S_n$.

Given a stream $\Sigma=\langle S_0,\dots, S_n\rangle$ and an \system program \mgp stratified by $\Pi_1 \dots\Pi_k$, let  $\Sigma^{\prime}$ be the stream $\Sigma^{\prime}=\langle S'_0, \dots, S'_{n-1}, S_n\rangle $ where $S'_0=$ ${\cal P}(\gp, \langle S_0\rangle)$ and each $S'_i$ for $i \in \{1,\dots,n-1\}$ is defined as ${\cal P}(\gp, \langle S'_0,\dots,S'_{i-1},S_i\rangle)$. We define ${\cal R}(\gp, \Sigma^{\prime})$ as \emph{streaming model of \mgp on $\Sigma$}.

The definition of the streaming model of an \system non-restricted program differs from the one given for restricted programs as it considers the persistent outcome over strata for time points up to $n-1$ and the outcome over strata for $n$ including heads of rules in the form $(2)$.
\begin{example}
Let $\Sigma$ be the stream of the Example\re{ex:output} and $\gp_3$ be the following program:
\begin{larscode}

\centering
$r_1$:\ \textbf{\#temp} c(X) \derives b(X). \ \ \ \ \ \ \ 
$r_2$:\ d(X) \derives c(X) \fin{1}.
\end{larscode}
Since the first rule of $\gp_3$ is of the form (2), we have that $S'_0=$ ${\cal P}(\gp_3, \langle S_0\rangle)= {\cal P}(\gp_3, \langle\{b(5)\}\rangle)= \{b(5),d(5)\}$ and $\Sigma^{\prime}=$ $ \langle S'_0, S_1\rangle$= $\langle\{b(5),d(5)\},\{c(7)\}\rangle$. The streaming model of $\gp_3$ on $\Sigma$ is the set ${\cal R}(\gp_3,\Sigma^{\prime})={\cal R}(\gp_3, \langle\{b(5),d(5)\},\{c(7)\}\rangle)=\{c(7),d(7)\}$.
\end{example}

\smallskip
\noindent\textbf{\textit{Modelling SR Problems.}}\label{subsec:KR}
We next show how practical problems \replace{requiring}{that require} reasoning over streams can be modelled via the \system language.
Note that, besides streaming literals, \system also supports built-\add{in} atoms and aggregate literals as defined in the ASP-Core-2 standard\cit{DBLP:journals/tplp/CalimeriFGIKKLM20}; currently, the only restriction is that aggregate elements cannot feature (non-degenerate) streaming literals.
For the sake of readability, we omitted their description in the language syntax; the following simple program briefly shows their usage: in a scenario where the total number of cars passed now or in the previous $20$ time points must be computed, rule $r_1$ counts the number of passes for each car, while $r_2$ uses an ASP aggregate to compute the total.
\begin{larscode}

$r_1:$ carPassing(C,N)\derives car(C) \countop{N}{20}.\ \\
$r_2:$ tot(T)\derives \#sum\{N,C: carPassing(C,N)\}=T.
\end{larscode}

\smallskip\noindent{\textit{Photo-voltaic system.}}\label{par:photo-voltaic-system}
Suppose that we need to build an Intelligent Monitoring System (IMS) for a photo-voltaic system (PVS) to promptly detect malfunctions.
Without going into technical details, for the sake of simplicity, let us suppose that the PVS is composed by a grid of interconnected panels via solar cables and each panel is provided with a sensor that measures the amount of energy produced and continuously sends data to the IMS.
Each panel continuously produces energy to be transferred to a Central Energy Accumulator (CEA), directly \replace{of}{or} via a path between neighbour panels across the grid.
Let us assume that a time point correspond\add{s} to a second.
A panel is working if it is known to have produced an amount of energy greater than a given threshold within the last \replace{$5$}{$4$} seconds, and, in addition, if it is reachable by the CEA (i.e., there exists a path of working panels linking it to the CEA).
If some unreachable working panels have been detected more than $2$ times in the last $3$ seconds, an alert must be raised for an identified malfunction.
Furthermore, the IMS must request a maintenance intervention if the failure is continuously observed for $5$ seconds.
This scenario can be modeled via an \system program as reported next.

\begin{larscode}
%
%

$r_1:$\  workingPanel(P) \derives energyDelivered(P,W) \atleast{1}{4}, \\
\hspace*{3.7cm} energyThreshold(Et), W>=Et.\\
$r_2:$\  reachable(cea,P2) \derives link(cea,P2), workingPanel(P2).\\
$r_3:$\  reachable(P1,P3) \derives reachable(P1,P2), link(P2,P3), workingPanel(P3).\\
$r_4:$\   unlinked \derives workingPanel(P), not reachable(cea,P).\\
$r_5:$\  regularFunctioning \derives unlinked \atmost{2}{3}.\\
$r_6:$\   alert \derives not regularFunctioning.\\
$r_7:$\  callMaintenance \derives alert \always{5}.

\end{larscode}

The predicates \mono{link} and \mono{energyThreshold} represent 
the PVS configuration and the threshold defining a working panel for the given reasoning interval; these data do not change during such interval.
The predicate \mono{energyDelivered} represents the amount of power produced by each panel; at each time point within the reasoning interval, the current values are sent to the IMS, thus producing a stream.
Rule $r_1$ defines a panel as working if it transmitted an amount of energy greater than the threshold at least once in the interval from the current time point to the previous 4 consecutive ones, i.e., the last $4$ seconds.
Rules $r_2$ and $r_3$ recursively define the set of reachable working panels starting from the CEA.
Rule $r_4$ detects if there are unlinked working panels and
$r_5$ defines proper functioning by checking that the atom \mono{unlinked} appeared no more than two times in the last 3 seconds. Eventually,
$r_6$ raises up an alert if there is not a regular functioning and
$r_7$ asks to call the maintenance if an alert has been raised in all the last $5$ seconds.





\smallskip\noindent{\textit{Underground Traffic Monitoring.}}
Let us imagine we want to build a monitoring system for the underground trains in the city of Milan.
In this example, we suppose that a time point correspond\add{s} to a minute; given a station, passengers expect to see a train stopping every $3$--$6$ minutes, during the rush hours.
The following \system program models a simple control system that warns passengers when this regularity is broken to several extents (i.e., mild/grave irregularity).
This would allow, e.g., to properly mark each station on a map of a mobile/web app.

\begin{larscode}

$r_1:$ irregular \ \derives train\_pass, train\_pass \replace{\countopd{X}{1,2}, X>0} {\atleastd{1}{1,2}}.\\
$r_2:$ irregular \ \derives not train\_pass \fin{6}. \\
$r_3:$ \textbf{\#temp} num\_anomalies(X) \derives irregular \countop{X}{30}.\\
$r_4:$ mild\_alert \ \ \derives num\_anomalies(X), X>2, X<=5.\\
$r_5:$ severe\_alert \derives num\_anomalies(X), X>5.
\end{larscode}

Rules $r_1$ and $r_2$ are used to detect irregularities: trains arriving too early or too late.
Rule $r_3$ counts the number of irregular situations in the last half an hour, producing an instance of the \mono{num\_anomalies}; $r_4$ and $r_5$ raise the proper warning.
Note that \mono{num\_anomalies} is intended to be an auxiliary predicate whose instance is used for determining irregularities only watching at the current time point; thus, it is marked as \textbf{\#temp}, so that its instance contributes to the current streaming model but it is no longer considered in next time points.



\section{\system Architecture}\label{sec:architecture}
The system takes as input an \system program $P$ and a stream $\Sigma=\langle S_0,\dots,S_n\rangle$ and iteratively builds a stream $O=\langle O_0,\dots,O_n\rangle$ such that each $O_t$ contains the result of the evaluation at the time point $t$.
Figure\re{fig:env_work-flow} depicts the high-level system architecture that consists of three main modules: \EM, \SM and \CM, all making an ad-hoc use of \flink APIs.
Details about each module are given below.

\begin{figure}[h]
\centering
  \subfloat[General Architecture.]{\includegraphics[width=5.8cm]{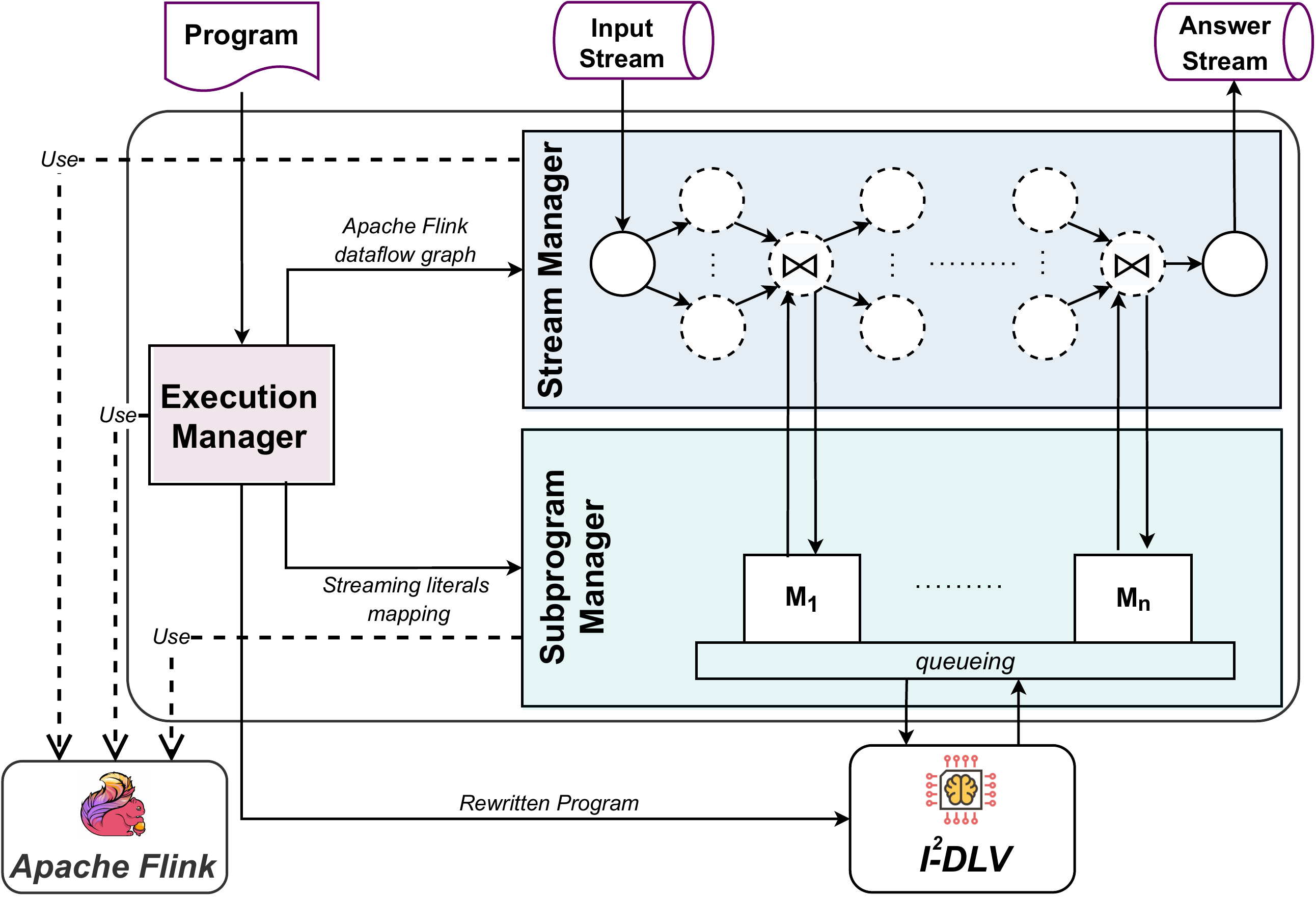}\label{fig1}}
  \hfill
  \subfloat[Architecture specialized for an example program. ]{\includegraphics[width=7.2cm]{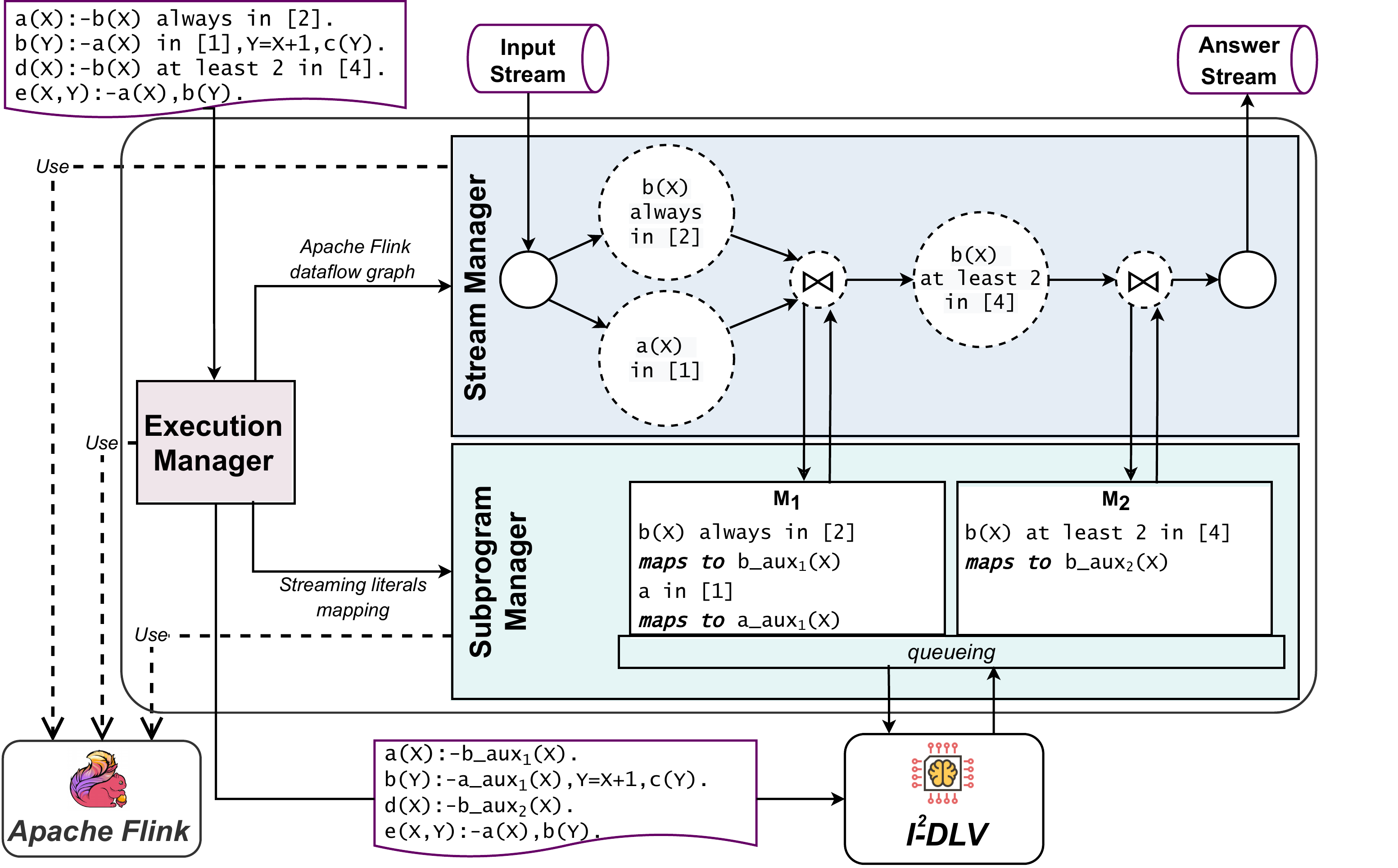}\label{fig2}}
   \caption{System Architecture.}\label{fig:env_work-flow}
 \vspace{-0.4cm}
\end{figure}

\subsection{\EM}\label{sub_sec:em}
The \EM is in charge of setting up the evaluation of an \system program \mgp.
First, it computes from \mgp a flat program $\gp'$ and determines a mapping \mapping from streaming atoms to predicate atoms.
Moreover, it divides \mgp into subprograms taking into account dependencies among all rules in \mgp caused by streaming atoms so that each subprogram can be separately processed, limiting the interplay between \flink and \iidlv.
On the basis of such program splitting, it constructs the \flink dataflow graph (cf. Section\re{sec:components}).
Finally, it provides: the \SM with the dataflow graph, the \CM with \mapping, and \iidlv with $\gp'$.
The \EM tasks are detailed below.

\smallskip
\noindent
\textbf{\textit{Program Rewriting.}}
The \EM produces a flat program $P'$ from $P$. Each streaming atom $p(t_1,\ldots,t_n)\; \mathbf{\diamond} \inm \{d_1,\dots, d_m\}$, with
$\diamond \in \{\al c$, $\am c$, $\alwm$, \textbf{count} $c\}$, which is not in the degenerate form and so that $c$ is a constant, is replaced by $p'(t_1,\ldots,t_n)$; each streaming atom $p(t_1,\ldots,t_n)$ \textbf{count} $X \inm \{d_1,\dots, d_m\}$ where $X$ is a variable, is replaced by $p'(t_1,\ldots,t_n,X)$; in both cases, $p'$ is a fresh predicate name.
Such replacements are stored in a mapping \mapping.
Ground instances of fresh predicates will be generated by the \CM on the basis of the evaluation of corresponding streaming atoms performed by the \SM.
\begin{example}\label{ex:program_rewriting}

Below are reported a program $\gp_4$ (left) and the flat program $\gp_4'$ obtained by rewriting it (right).

\noindent\begin{minipage}[h]{0.51\textwidth}
\begin{larscode}

$r_1:$ a(X) \derives b(X) \always{2}.\\
$r_2:$ b(Y) \derives a(X) \fin{1}, Y=X+1, c(Y).\\
$r_3:$ d(X) \derives b(X) \atleast{2}{4}.\\
$r_4:$ e(X,Y) \derives a(X), b(Y).
\end{larscode}
\end{minipage}
\begin{minipage}[h]{0.49\textwidth}
\begin{larscode}

$r_1':$ a(X) \derives b$\_$aux1(X).\\
$r_2':$ b(Y) \derives a$\_$aux1(X), Y=X+1, c(Y).\\
$r_3':$ d(X) \derives b$\_$aux2(X).\\
$r_4:$ e(X,Y) \derives a(X), b(Y).
\end{larscode}
\end{minipage}
The mapping of the replacements is \mapping{\small{$=\{$\mono{b(X) \always{2}} $\mapsto$ \mono{b\_aux1(X)}, \mono{a(X) \fin{1}} $\mapsto$ \mono{a\_aux1(X)}, \mono{ b(X) \atleast{2}{4}} $\mapsto$ \mono{b\_aux2(X)} $\}$}}.
\end{example}

\smallskip\noindent\textbf{\textit{Program Splitting and Processing Order.}}\label{par:modularization}
Given an \system program \mgp, we build a directed labeled graph called \textit{Stream Dependency Graph} and denoted $\sdg$, whose nodes are the predicates in rule heads of \mgp and for each pair of nodes $p$ and $q$, there is an arc ($p$, $q$) if there exists a rule $r \in \gp$ such that $preds(H(r)) = \{q\}$ and $p \in preds(B(r))$.
In case $p$ occurs in a streaming literal that is not in the degenerate form, then the arc is labeled with $\quot{<}$; no label is added otherwise.
On the basis of \msdg, we define an additional directed labeled graph, called \textit{Stream Component Graph} and denoted $\scg$: its nodes are the strongly connected components of \msdg (i.e., sets of predicates), and there is an arc from component $A$ to component $B$ if there exists an arc ($p$, $q$) in \msdg such that $p \in A$ and $q \in B$.
Each arc ($A$, $B$) in $\scg$ is labelled with $\quot{<}$ if there exists at least one arc ($p$, $q$) labeled with $\quot{<}$ in \msdg such that $p\in A$ and $q\in B$; no label is added otherwise.

Relying on \mscg, rules of \mgp that can be processed together are grouped, and a processing ordering among groups of rules is established.
For any pair of nodes $A$ and $B$ of \mscg, we say that $A$ \textit{precedes} $B$ (denoted $A \prec B$) if there exists a path in \mscg from $A$ to $B$ containing at least one arc labeled with $\quot{<}$; we say $A$ \textit{is alongside} $B$ (denoted $A \approx B$) otherwise.
We identify an ordering $C_1,\dots,C_n$ of the nodes of \mscg such that, for each $i<j$, it does not hold that $C_j\prec C_i$.
According to such an ordering $C_1,\dots,C_n$, \remove{under certain conditions,} we \remove{can} collect the predicates of some consecutive nodes in order to form a macro-node \replace{; in particular,}{as follows:} for each pair of nodes $C_i$ and $C_k$, with $i\le k \le n$ we \remove{can} construct the macro-node $\mmod{}=\bigcup_{j} C_j$ with $i \le j \le k$ such that either $i=k$ (i.e., the macro-node actually consists of a single node), or for each $j\neq k$, $C_j \approx C_{j+1}$ (i.e., the macro-node consists of nodes that are all alongside each other).
We can therefore define an ordered sequence of all the maximal macro-nodes such that for any pair of macro-nodes $M_1 = \bigcup_{j_1} C_{j_1}$ with $i_1 \le j_1 \le k_1 \le n$ and $M_2 = \bigcup_{j_2} C_{j_2}$ with $i_2 \le j_2 \le k_2$, if $M_1$ precedes $M_2$ in the sequence, then $k_1<i_2$.
This ordered sequence induces both a splitting of \mgp into subprograms and a processing order for them: for each macro-node \mmmod{}, the subprogram $\gp_\mmod{}$ is the set of all rules $r$ of \mgp such that such that $preds(H(r))\subseteq \mmod{}$, i.e., the predicate occurring in $H(r)$ belongs to \mmmod{};
the processing order for the subprograms coincides with the ordering in the sequence of the corresponding macro-nodes.
Note that for an \system program several orderings of the nodes of \mscg might exist, in general and therefore, different processing orders might be obtained. In the following, we will refer to one of these processing orders as $Ord_\gp$.

Eventually, a rule $r$ within a \replace{sub-program}{subprogram} $\gp_\mmod{}$ is \emph{streaming-recursive} if there is a cycle in the Stream Dependency Graph \msdg among two nodes $p$ and $q$ where $p\in preds(B(r))$ and $q\in preds(H(r))$ such that there is at least an arc labeled with $\quot{<}$.

\begin{example}\label{ex:module_example}
\begin{figure}
\centering
         \includegraphics[width=0.5\textwidth]{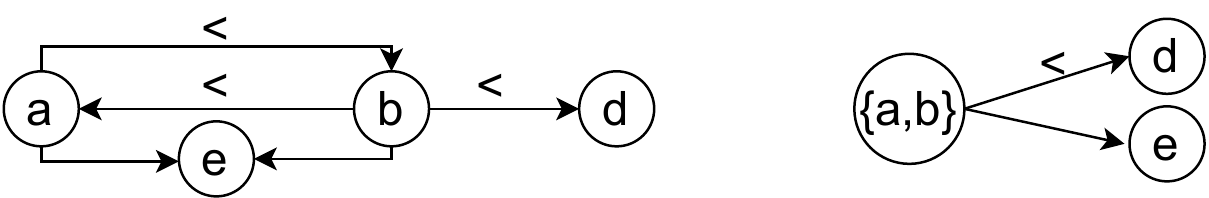}%
        \caption{Stream Dependency (left) and Component Graphs (right) of Example\re{ex:program_rewriting}.}
        \label{fig:graphs}
        \vspace{-0.4cm}
\end{figure}
Let us consider the program $\gp_4$ of Example\re{ex:program_rewriting}. Figure\re{fig:graphs} shows the Stream Dependency and Component Graphs of $\gp_4$. We can observe that $G_{\gp_4}^{SD}$ has three strongly connected components: $\{a,b\}$, $\{d\}$ and $\{e\}$.
According to $G_{\gp_4}^{SC}$, we have: $\{a,b\}\prec \{d\}$, $\{a,b\}\approx \{e\}$,$\{e\}\approx\{a,b\}$, $\{d\}\approx \{e\}$ and $\{e\}\approx\{d\}$.
As a consequence, there are four orderings:
        $o_1=\{\{e\},\{a,b\},\{d\}\}$;
        $o_2=\{\{a,b\},\{e\},\{d\}\}$;
        $o_3=\{\{a,b\},\{d\},\{e\}\}$;
        $o_4=\{\{a,b\},\{e\},\{d\}\}$.
%
Let us focus on $o_1$. The maximal macro-nodes that we can build are $\{a,b,e\}$ and $\{d\}$ and the resulting program splitting is: $\gp_4{_{\{a,b,e\}}}=\{r_1, r_2, r_4\}$,
$\gp_4{_{\{d\}}}=\{r_3\}$. Moreover, $r_1$ and $r_2$ in $\gp_4{_{\{a,b,e\}}}$ are streaming-recursive. Note that, $o_2$ would induce the same maximal macro-nodes and in turn, the same splitting of $o_1$ and in both cases, $\{r_1, r_2, r_4\}$ are processed before $\{r_3\}$.
\end{example}

\smallskip\noindent\textbf{\textit{\flink dataflow graph creation.}}
As described above, given an \system program, the \EM splits it into subprograms and computes a processing order $Ord_\gp$.
Then, according to $Ord_\gp$, it builds an \flink dataflow graph by adding one or more nodes for each subprogram and for each of its streaming atoms.
As shown in Figure\re{fig:env_work-flow}, there are two kinds of nodes represented as empty or $\Join$-filled dashed circles.
Empty circles correspond to operators needed for evaluating a streaming atom; \replace{we purposely designed some operators for each type of streaming atom, implementing its corresponding semantics, for determining whether it is entailed.}
{by means of \flink APIs, we defined some operators in order to implement the semantics of \textit{in}, \textit{always}, \textit{count}, \textit{at least}, and \textit{at most}.}
$\Join$-filled circles correspond to \add{custom} operators intended to join results of the evaluation of all streaming atoms in a subprogram and to set up the evaluation of such subprogram. \add{We implemented within such operators, the low-level transformations needed for filtering and joining outcomes of linked circles.}
Empty circles are connected with directed arcs to the $\Join$-filled one associated to the subprogram in which the corresponding streaming atoms occur.
$\Join$-filled circles are linked to the empty ones associated to the streaming atoms appearing in the subsequent subprogram in $Ord_\gp$.

\subsection{\SM and \CM}\label{sub_sec:sm_cm}
As depicted in Figure\re{fig:env_work-flow}, the \SM executes the scheduled operators in the \flink dataflow graph and interacts with the \CM that, in turn, mediates communications with \iidlv responsible for the evaluation of the flat program $P'$, which in fact, is always a normal and stratified w.r.t. negation ASP program.
%
Let us assume to be at the time point $n$. The \SM handles the input stream $\Sigma=\langle S_0,\dots,S_n\rangle$ and the stream $O=\langle O_0,\dots,O_n\rangle$ where $O_0, \dots, O_{n-1}$ collect the results obtained from the evaluations in previous time points and $O_n$, initially empty, is filled in while subprograms are evaluated.

More in detail, let $\gp_\mmod{}$ be the first subprogram in $Ord_\gp$.
The \SM executes the operators in the dataflow graph corresponding to the streaming atoms in $\gp_\mmod{}$ over $\Sigma$ and $O$.
Then, it passes to the \CM $S_n$, $O_n$ and the set of holding streaming atoms $H$, i.e., those evaluated as true, at the time point $n$.
The \CM:
($i$) according to the mapping \mapping, generates from $H$ the set $H'$ of holding ground instances of the fresh predicates introduced by the \EM;
($ii$) provides $S_n\cup O_n\cup H'$ as input to \iidlv that incrementally evaluates $P'$ over such input and computes the corresponding unique answer set $A$;
($iii$) receives $A$ (i.e., a set of ground (predicate) atoms) and adds $A$ to $O_n$.
The \SM takes control back, evaluates the next nodes in the dataflow graph and interacting with the \CM, performs the evaluation of the next subprogram \remove{$\gp_\mmod{}$} in $Ord_\gp$.
The process continues until all subprograms are evaluated according to $Ord_\gp$ \replace{and}{;} the output at $n$ is $S_n\cup O_n$.
Note that the smaller it is the number of subprograms, the smaller it is the number of iterations between \iidlv and \flink: indeed, maximal macro-nodes \replace{permit to limit}{are limiting} such interplay.

In case a subprogram contains some streaming-recursive rules $R$, the \SM has to repeatedly evaluate the streaming atoms in $R$, i.e., every time $O_n$ is enriched by the \CM. When no more new ground (predicate) atoms can be derived for the predicates in the heads of $R$, the above described process goes on with the next subprogram in $Ord_P$.

\add{
Note that, the set $S_n\cup O_n$ actually coincides with the streaming model of \mgp on $\Sigma$. Indeed, the streaming model is independent from the stratification of choice (see Section\re{sec:language}); thus, if we consider the stratification where each stratum is the smallest possible, then each subprogram consists of rules that might belong to one or more consecutive strata; when evaluating a subprogram $\gp_\mmod{}$,
\iidlv by design takes into account dependencies among the strata and processes rules in $\gp_\mmod{}$ accordingly.
Thus, the output computed for $\gp_\mmod{}$ coincides with the output that we would obtain when evaluating, one after the other, the strata contained in $\gp_\mmod{}$ as well as in all subprograms preceding $\gp_\mmod{}$, according to the definition of streaming model.
In other words, the output computed for $\gp_\mmod{}$ coincides with the streaming model of the \system program constituted by all subprograms up to $\gp_\mmod{}$ on $\Sigma$. It is worth noting that the \SM implements a queuing mechanism for the subtasks that have to be performed for evaluating $P$ on $\Sigma$ exploiting \flink APIs.
Thanks to this, \system is able to manage backpressure\cit{DBLP:conf/icde/KarimovRKSHM18} guaranteeing no loss of data, and thus the correctness of the output.
}

\begin{example}
\label{ex:prew_example}
Consider again $\gp_4$ of Example\re{ex:program_rewriting}. Figure\re{fig:env_work-flow} $(b)$ illustrates the evaluation process of $\gp_4$
by referring to the general architecture.
We assume to be at the $n$-th time point and that $\Sigma=\langle S_0,\dots,S_n\rangle$ is the input stream, while $O=\langle O_0,\dots,O_n\rangle$ with $O_n=\emptyset$ is the stream iteratively built so far.

The \EM provides \iidlv with the flat program $\gp'_4$ obtained by rewriting $\gp_4$, splits $\gp_4$ and identifies the processing order for its subprograms.
Suppose as shown in Example\re{ex:module_example} that the splitting is $\gp_4{_{\{a,b,e\}}}= \{r_1, r_2, r_4\}$ and $\gp_4{_{\{d\}}}=\{r_3\}$ and that $\gp_4{_{\{a,b,e\}}}$ precedes $\gp_4{_{\{d\}}}$.
The \EM accordingly creates the dataflow graph and passes it to the \SM.

The dataflow graph contains three empty dashed circles for the operators needed for evaluating streaming atoms in $\gp_4{_{\{a,b,e\}}}$ and $\gp_4{_{\{d\}}}$, and two $\Join$-filled ones that receive results from the linked empty circles and forward them to \CM.
More in detail, the \SM first evaluates the empty dashed circles relative to $\gp_4{_{\{a,b,e\}}}$ and then, the \CM is required to take into account $\gp_4{_{\{a,b,e\}}}$ hence, producing the ground instances for $\mathtt{a\_aux1}$ and $\mathtt{b\_aux1}$ and properly invoking \iidlv.
The result received by the \CM is forwarded back to the \SM that updates $O_n$. Such loop between \SM and \CM for processing of $\gp_4{_{\{a,b,e\}}}$ continues until nothing new can be inferred for the predicates $\{a,b,e\}$ and only after $\gp_4{_{\{d\}}}$ is processed.
\end{example}

\section{Experimental Evaluation}\label{sec:experiments}
In order to assess reliability and performance of \system, we carried out an experimental activity over different SR problems and settings.
All experiments have been performed on a NUMA machine equipped with two $2.8$GHz AMD Opteron 6320 CPUs, with 16 cores and $128$GB of RAM.
For the sake of reproducibility, both the system and the whole set of experiments are available at \url{https://demacs-unical.github.io/I-DLV-sr}.

We conducted two kind of analysis.
First, we wanted to compare the performance of \system with some other available logic-based stream reasoner.
A number of implementations are available (see Section~\ref{sec:related_works}); however, they considerably differ in syntax and/or semantics, and also from an architectural/implementative point of view; thus, a fair comparison is rather difficult.
Given the distributed nature of \system, we compared it to Distributed-SR\cit{eiter2019distributed}, which is the most recent LARS-based implementation, supports a large set of features and relies on a distributed architecture. The latest available version has been executed.
For the former analysis, we considered two benchmarks: \mmedia and \hjoin.
\mmedia~\cite{DBLP:conf/icc/BeckBDEHS17,eiter2019distributed} is a real-world benchmark that requires to manage the caching policy of a video content over an incoming stream that describes the evolving popularity level of the content.
Besides the original problem~\cite{DBLP:conf/icc/BeckBDEHS17}, we considered a slightly different version that deals with more than one event per time point (we refer to a true atom at a time point in the stream as an ``event''). 
Therefore, the encoding is adapted to handle more than one video content, and the incoming stream contains a number of instances, representing the popularity levels for each content, ranging from $50$ to $500$.
\hjoin is an artificial problem conceived in order to test scalability.
It consists of the single rule: \mono{a(X,Y):-b(X,Z) in [w],c(Z,Y) in [w].} where $w$ is $2$ or $20$, depending on the experiment.
The input streams feature an equal number of instances of predicates $b$ and $c$ whose total ranges from $50$ to $500$.
In addition, in order to test \system specific features, we conducted a performance analysis on the problem introduced in Section\re{par:photo-voltaic-system}, namely \pvsystem, over grids of increasing size ranging from $20 \times 20$ panels with $11,970$ links up to $30 \times 30$ panels with $60,682$ links. \add{In this data-intensive domain, we performed a further analysis to check the advantages of incrementality on \system performance.}

\smallskip

\begin{figure}[t]
    \includegraphics[width=0.8\textwidth]{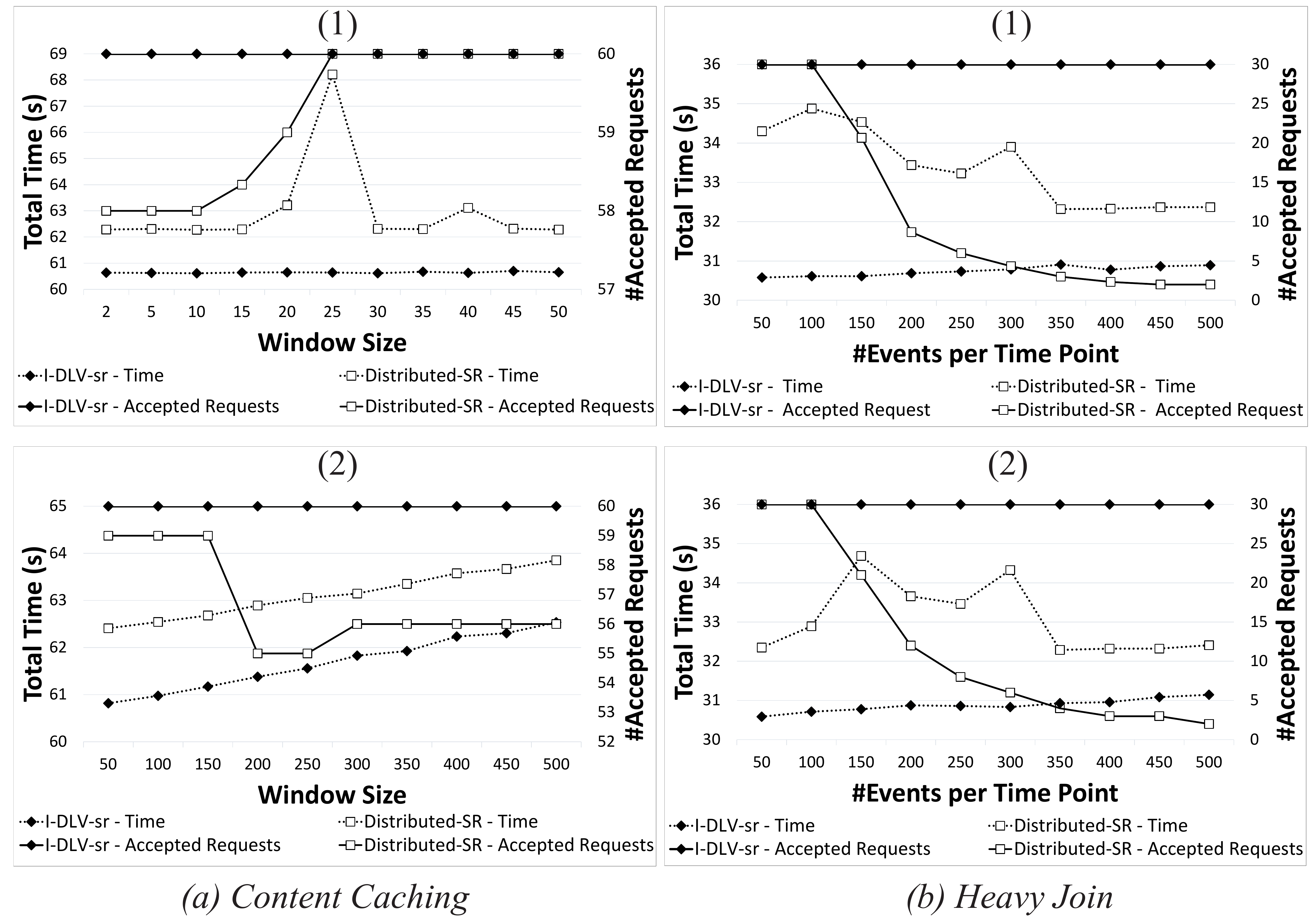}
    \caption{\replace{Experimental Results}{Results on \mmedia and \hjoin.}}
    \label{fig:ticker}
    \vspace{-0.4cm}
\end{figure}

Results are reported in Figures~\ref{fig:ticker} \add{ and\re{fig:pvsystem}}; in each plot,  a marker in a line corresponds to the average of the results of three executions; in each execution, the tested system received inputs for $60$ time points for \mmedia and \pvsystem and $30$ for \hjoin.
In the comparison with Distributed-SR, input arrives at a frequency fixed to 1 time point per second, while in the \pvsystem experiments, frequency varies.
Our analysis focuses on three measures: \time, \req and \lat:
\time represents the total elapsed time for each execution, excepting the initial time spent for the set up;
\req is the number of accepted incoming requests computed by checking the system logs, and counting the number of time points whose corresponding input is read by the system;
\lat is the processing-time latency, i.e., the interval between the time at which the system receives the input relative to a time point and the time at which the system returns the corresponding output.

The left column of Figure~\ref{fig:ticker} refers to \mmedia tests.
Plot~($a.1$) is related to experiments on the original encoding, where each time point features a single incoming event; it reports \time (left $y$-axis) and \req (right $y$-axis), while window sizes vary from $2$s to $50$s ($x$-axis).
The two systems appear to have similar behaviours w.r.t. \time, slightly above $60$s on average, that is the minimum amount of time needed for waiting all the incoming $60$ requests (recall that here the frequency is of a time point per second).
The picture changes when looking at the right $y$-axis: while \system accepts and correctly returns the expected output for all the $60$ requests, Distributed-SR fails in handling them all.
Plot~($a.2$) refers to the modified encoding mentioned above, with window sizes set to $5$ seconds.
For each execution, the graph plots the two measures considered above; on the $x$-axis, the number of events per time point \remove{varies} ranges from $50$ to $500$.
The two systems show similar trends in total times, that grow along with the number of events for time points.
As in the previous case, a loss of incoming requests is reported for Distributed-SR, becoming more evident for greater numbers of events.

Results for \hjoin are displayed in the \replace{central}{right} column of Figure\re{fig:ticker}; the analysis is the same of Plot~($a.2$).
We considered window size of $2$s in Plots~($b.1$) and $20$s in ($b.2$), respectively.
The two plots depict similar results.
In terms of \time, it is more evident w.r.t. the \mmedia case that \system performs better than Distributed-SR, and it maintains the same trend in both cases, while Distributed-SR exhibits a fluctuating behaviour.
As for \req, Distributed-SR ``misses'' a relevant number thereof: the loss is limited when the number of events per time point is small, but rapidly grows with the number of events for time point, so that in the end almost all incoming requests are ignored.

\begin{figure}[t]
    \includegraphics[width=0.8\textwidth]{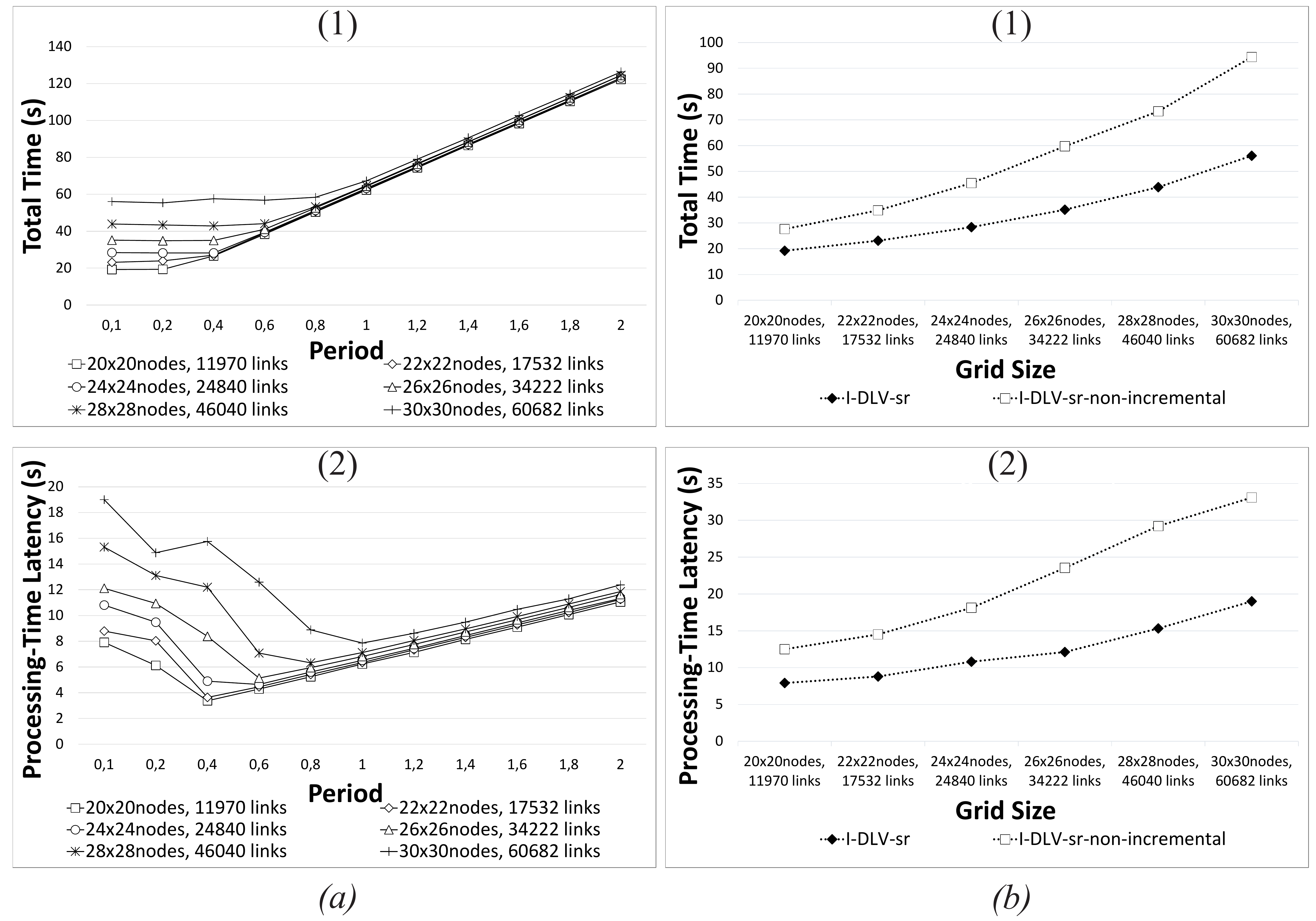}
    \caption{Results on \pvsystem.}
    \label{fig:pvsystem}
    \vspace{-0.4cm}
\end{figure}

\replace{The right column of Figure\re{fig:ticker}}{The left column of Figure\re{fig:pvsystem}} depicts \add{the} results on the \pvsystem benchmark where \system has been tested over different grids with increasing sizes, one size per line. This is a quite expensive domain as rules at lines $2$--$3$ encode a reachability task among working grid panels that vary over time.
Plots\replace{~($c.1$) and~($c.2$)}{~($a.1$) and~($a.2$)} respectively report \time and \lat when on the $x$-axis, the period of incoming requests varies ranging from $0.1$s to $2$s.
In this case, \req is not reported as the system correctly processes all the $60$ requests.
As for \time, the six lines in the plot show almost the same trend: constant up to a certain period $P$, and then linearly growing.
Note that, for greater grid sizes, also $P$ is greater: this is expected, and can be explained by observing \lat in Plot~\replace{($c.2$)}{($a.2$)}.
Indeed, for each line in the plot, \lat is smaller when the period is close to its $P$ value beyond which we observe the  so-called sustainable throughput\cit{DBLP:conf/icde/KarimovRKSHM18}.
This is because for period values smaller than $P$, \system is asked to process requests that come more often, so it starts to enqueue pending requests.
When the period is greater than $P$, instead, \system is able to consider a request as soon as it arrives, thus no queuing is needed, and the greater the period is w.r.t. $P$, the greater it is the idle time between the complete processing of a request and the incoming of the next one.
Periods close to $P$ are ideal, as idle time is close to $0$ and no queuing occurs.
\add{
The right column of Figure\re{fig:pvsystem} compares the performance of two different versions of \system versions: one relies on the incremental \iidlv system, whilst the other relies on the non-incremental \idlv engine (v. 1.1.6).
Plots~($b.1$) and~($b.2$) report \time and \lat, respectively, of the two tested versions of \system, when on the $x$-axis the grid sizes vary; the period of incoming requests is set to $0.1$s.
We observe that the version relying on the incremental evaluator keeps \time and \lat smaller than the other; the gap between the two lines becomes more and more evident as the grid size increases, suggesting that \system scales better when relying on the incremental-based version.
Intuitively, the gain is attributable to the incremental computation of the transitive closure of the reachable relation over working panels; indeed, the adopted overgrounding technique maintains and reuses all the previously computed ones, rather than recomputing from scratch all the connections.
We conclude by mentioning that the same analysis about incrementality has been also done for \mmedia and \hjoin; however, as these domains are not data-intensive, the evaluation time per rule is so small that recomputing everything from scratch each time is not that expensive. Results  (see \re{sec:appendixB}) show that the incremental-based version behaves as the non-incremental one without suffering from potential overheads.
}



\section{Related Works}\label{sec:related_works}
Stream Reasoning has been subject of a number of researches over the latest years.
Nevertheless, there are no standardized formalisms nor techniques for SR to date, making the comparison among approaches relying on different semantics and technologies rather difficult.
Apart several relevant solutions stemmed in the semantic-web context~\cite{DBLP:journals/ijsc/BarbieriBCVG10,DBLP:conf/semweb/PhuocDPH11,DBLP:conf/lpnmr/PhamAM19,hoeksema2011high}, the proposals that relate most with ours are the ones based on ASP.
With this respect, one well-established is LARS: Logic for Analytic Reasoning over Streams\cit{beck2018lars}, a formal framework enriching ASP with temporal modalities and window operators.
LARS theoretically consists of a full-fledged non-monotonic formalism for reasoning over streams; indeed, the full language is computationally intractable.
Current implementations, such as Laser\cit{bazoobandi2017expressive}, Ticker\cit{beck2018ticker}, and its recent distributed version\cit{eiter2019distributed}, support smaller, yet practically relevant fragments.
For performance reasons, Laser handles only negation-stratified and stream-stratified programs, i.e., recursion is not supported if \add{it} involves negation or windows\cit{beck2015answer}.
Ticker comes with two evaluation modes: one makes use of the state-of-the-art ASP system clingo~\cite{DBLP:journals/tplp/GebserKKS19} as back-end, and is intended for stratified programs, i.e., programs having only a single model; the other mode uses incremental truth maintenance techniques under ASP semantics, and, in case of multiple solutions, computes and maintains one single model randomly chosen.
Distributed-SR is an additional version of Ticker, recently released; in order to increase the throughput, it implements an interval-based semantics of LARS that relies on Ticker as internal engine; it supports distributed computation, at the price of disabling the support for recursion through window operators.
All Ticker versions require any variable appearing in the scope of a window atom to be ``guarded'' by some standard atom including it; such variables are grounded upfront in a so-called pre-grounding phase.

\system supports the ASP fragment stratified w.r.t. negation, which is extended with streaming literals over temporal intervals: recursion involving streaming literals is freely allowed as well.
The implementation is designed for supporting incremental evaluation of \system programs, thanks to the integration with \iidlv, and parallel/distributed computation, thanks to the integration with \flink.
From the one hand, incrementality allows to efficiently evaluate 
logic programs at each time point and to avoid the need for pre-grounding; also, apart from ``canonical'' safety, no restriction is required for variables in streaming literals.
On the other hand, parallelism and distribution pave the way to the efficient evaluation of streaming atoms.
During execution, each operator in the \flink dataflow can have one or more operator subtasks, that are independent from each other; hence, they can be executed in different threads and possibly on different machines or containers.
In case of recursive subprograms, distribution is limited by evaluating the involved streaming atoms via single-threaded,  undistributed operators.


\replace{A logic-based SR system appearing to have an architectural approach close to the one herein proposed is BigSR\cit{DBLP:conf/bigdataconf/RenCNX18}, a distributed stream reasoner released in two implementations
built on top of the state-of-the-art stream processors Spark Streaming(\url{https://spark.apache.org}) and Flink, respectively.
The input language is a fragment of LARS that features
a limited set of window operators and does not include negation. For computing the semantics
BigSR relies on ad-hoc algorithms, rather than properly interacting with a logic-based system.}
{A distributed stream reasoner that has in common with \system the usage of a stream processor is  BigSR\cit{DBLP:conf/bigdataconf/RenCNX18}; it is released in two versions built on top of the state-of-the-art stream processors Spark Streaming (\url{https://spark.apache.org}) and \flink, respectively.
BigSR implements internal reasoning algorithms to compute the semantics, and significantly differs from \system on the supported language, the input format and the provided features.
In particular, it is oriented towards ontology-based reasoning and accepts RDF input streams.
Moreover, input programs fall in the positive fragment of plain LARS and can only contain the \textit{in} window operator.
In addition, depending on the stream processor, further limitations are required.
The version relying on Spark Streaming accepts stratified programs that can be recursive but have only global windows, i.e., the rules must share the same window operator.
The version built on top of \flink, instead, accepts non-recursive stratified programs, but with global windows at rule scope, i.e., all literals in a rule must share the same window operator.
In this latter version, such limitations are due to the BigSR usage of the multi-core/distributed nature of \flink, that makes the handling of synchronization of clocks, task progress and window trigger mechanisms difficult.
Furthermore, \system and BigSR differs for the adopted notion of time,
as the former relies on \textit{event time} notion
while the latter on  the \textit{ingestion time} one~\cite{hueske2019stream}.}

\add{When designing \system, Spark Streaming and Apache Storm\cit{DBLP:conf/sigmod/ToshniwalTSRPKJGFDBMR14} were evaluated as alternative stream processors.
Spark Streaming was excluded as, when receiving live input streams, it divides the data into batches, which are lately processed by the Spark engine to generate the final stream in batches.
On the other hand, \flink does not require the a-priori creation of batches and works in real-time, record by record, rather than batch by batch.
Apache Storm can handle data processing record by record
but, it does not provide the event time processing 
and does not natively offers the exactly-once semantics as instead \flink does.
This semantics ensures to \system that each incoming event affects the final outcome exactly once and, even in case of a machine or software failure, there is no data duplication nor unprocessed data.}

\section{Conclusions}\label{sec:future_work}

We presented \system, an ASP-based stream reasoner relying on a tight interaction between \iidlv and a \flink application, that in the \remove{first} experiments showed good performance and scalability.
\system is easily extendable by design; hence, we plan to add the support to additional language constructs while extending tests over new real-world domains.
Furthermore, we plan to study proper means to extend the language for the management of {\em noise} and {\em incompleteness}, and further move towards a more complete SR reasoner\cit{DBLP:journals/datasci/DellAglioVHB17}.

\newpage
\add{
\appendix
\section{}\label{sec:appendixA}
\paragraph{Proof of Proposition\re{prop:stratum}.}\ We first recall that by definition of stratum application on $\Sigma$, $\Sigma_0=\Sigma'_0=\Sigma$ and that a trigger application adds ground (predicate) atoms only to the last set of a stream.
For convenience, given a stream $\Sigma=\langle S_0,\dots, S_n\rangle$, we denote the last set $S_n$ as $last(\Sigma)$.
We prove that if an atom $a$ belongs to the last set of one of the two streams $\Sigma_h$ or $\Sigma'_t$, then $a$ necessarily belongs to the last set of the other stream.
Hence, let us suppose that $a\in last(\Sigma_h)$, we prove that $a\in last(\Sigma'_t)$. In particular, we show that $\forall i\in \{0,\dots,h\}, a\in last(\Sigma_i) \implies \exists j_a \in \{0,\dots,t\}: a \in last(\Sigma'_{j_a})$.
We proceed by induction:
\begin{compactitem}
\item[-] $a \in last(\Sigma_0)$. Since $\Sigma_0=\Sigma'_0=\Sigma$, we have that $j_a=0$.
\item[-] We assume that $a\in last(\Sigma_n) \implies \exists j_a\in \{0,\dots,t\}: a \in last(\Sigma'_{j_a})$.

\item[-] If $a\in last(\Sigma_{n+1})$ we can have that either $a\in last(\Sigma_{n})$ and by inductive hypothesis there exists  $ j_a\in \{0,\dots,t\}: a \in last(\Sigma'_{j_a})$ or $\Sigma_{n+1}$ is the result of the application of the trigger $\langle r_n, \sigma_n \rangle$ on $\Sigma_n$, i.e. $\Sigma_n\langle r_n, \sigma_n \rangle \Sigma_{n+1}$, with $a=\sigma_n(l)$ where $l\in H(r_n)$.
In the latter case, we have that $\Sigma_n \models \sigma_n(b) \forall b \in B(r_n)$.
If $b\in B(r_n)$ is a non-harmless literal its truth value cannot depend on rules belonging to stratum $\Pi_s$. Then $\Sigma_n \models \sigma_n(b)$ iff $\Sigma\models \sigma_n(b)$. If $b\in B(r_n)$ is an harmless literal with predicate atom $p(t_1,\ldots,t_p)$ we can have that $\Sigma \models \sigma_n(b)$ or we can have that $\sigma_n(p(t_1,\ldots,t_p)) \in last(\Sigma_n)$. By inductive hypothesis we have that
$\exists j_{b}\in \{0,\dots,t\}:\sigma_n(p(t_1,\ldots,t_p)) \in \Sigma'_{j_{b}}$. Hence, there exists a stream $\Sigma'_m$, with $m \in \{0,\dots,t-1\}$ such that $\Sigma'_m\models \sigma_n(b) \forall b \in B(r_n)$
and there exists $m < j_a\leq t$ such that $a\in \Sigma'_{j_a}$.

\end{compactitem}
\finex
}

\newpage
\section{}\label{sec:appendixB}
\begin{figure}[h]
    \includegraphics[width=\textwidth]{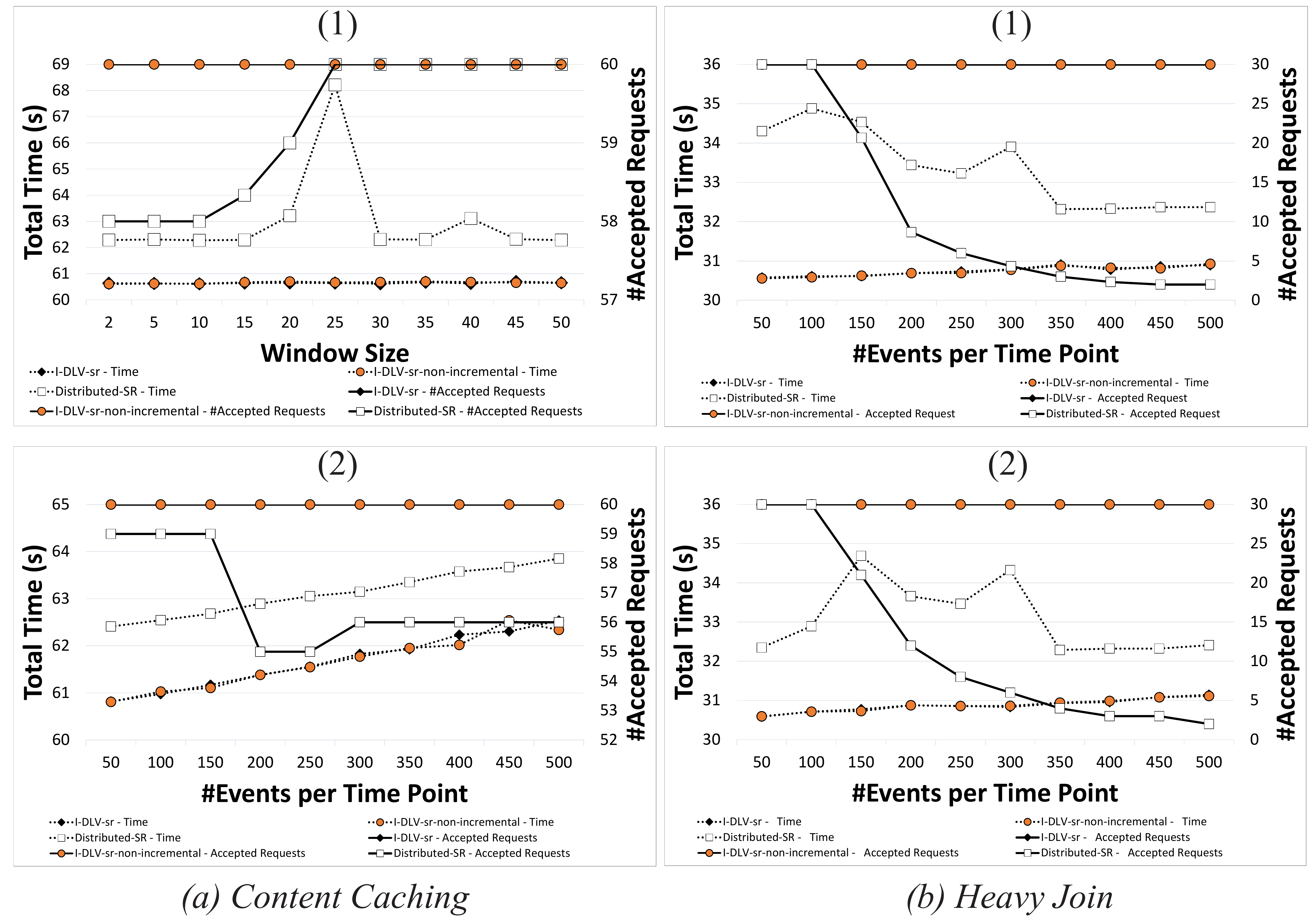}
    \caption{Results on \mmedia and \hjoin including also a version of \system relying on the non-incremental \idlv reasoner.}
    \label{fig:ticker_rew}
    \vspace{-0.35cm}
\end{figure}



%
%
%
\bibliographystyle{acmtrans}

\begin{thebibliography}{}

\bibitem[\protect\citeauthoryear{Barbieri, Braga, Ceri, Valle, and
  Grossniklaus}{Barbieri
  et~al\mbox{.}}{2010}]{DBLP:journals/ijsc/BarbieriBCVG10}
{\sc Barbieri, D.~F.}, {\sc Braga, D.}, {\sc Ceri, S.}, {\sc Valle, E.~D.},
  {\sc and} {\sc Grossniklaus, M.} 2010.
\newblock {C-SPARQL:} a continuous query language for {RDF} data streams.
\newblock {\em Int. J. Semantic Comput.\/}~{\em 4,\/}~1, 3--25.

\bibitem[\protect\citeauthoryear{Bazoobandi, Beck, and Urbani}{Bazoobandi
  et~al\mbox{.}}{2017}]{bazoobandi2017expressive}
{\sc Bazoobandi, H.~R.}, {\sc Beck, H.}, {\sc and} {\sc Urbani, J.} 2017.
\newblock Expressive stream reasoning with laser.
\newblock In {\em International Semantic Web Conference {(1)}}. LNCS, vol.
  10587. Springer, 87--103.

\bibitem[\protect\citeauthoryear{Beck, Bierbaumer, Dao{-}Tran, Eiter,
  Hellwagner, and Schekotihin}{Beck
  et~al\mbox{.}}{2017}]{DBLP:conf/icc/BeckBDEHS17}
{\sc Beck, H.}, {\sc Bierbaumer, B.}, {\sc Dao{-}Tran, M.}, {\sc Eiter, T.},
  {\sc Hellwagner, H.}, {\sc and} {\sc Schekotihin, K.} 2017.
\newblock Stream reasoning-based control of caching strategies in {CCN}
  routers.
\newblock In {\em {IEEE} International Conference on Communications, {ICC}
  2017, Paris, France, May 21-25, 2017}. {IEEE}, 1--6.

\bibitem[\protect\citeauthoryear{Beck, Dao{-}Tran, and Eiter}{Beck
  et~al\mbox{.}}{2015}]{beck2015answer}
{\sc Beck, H.}, {\sc Dao{-}Tran, M.}, {\sc and} {\sc Eiter, T.} 2015.
\newblock Answer update for rule-based stream reasoning.
\newblock In {\em {IJCAI}}. {AAAI} Press, 2741--2747.

\bibitem[\protect\citeauthoryear{Beck, Dao{-}Tran, and Eiter}{Beck
  et~al\mbox{.}}{2018}]{beck2018lars}
{\sc Beck, H.}, {\sc Dao{-}Tran, M.}, {\sc and} {\sc Eiter, T.} 2018.
\newblock {LARS:} {A} logic-based framework for analytic reasoning over
  streams.
\newblock {\em Artif. Intell.\/}~{\em 261}, 16--70.

\bibitem[\protect\citeauthoryear{Beck, Dao{-}Tran, Eiter, and Folie}{Beck
  et~al\mbox{.}}{2018}]{beck2018ticker}
{\sc Beck, H.}, {\sc Dao{-}Tran, M.}, {\sc Eiter, T.}, {\sc and} {\sc Folie,
  C.} 2018.
\newblock Stream reasoning with {LARS}.
\newblock {\em K{\"{u}}nstliche Intell.\/}~{\em 32,\/}~2-3, 193--195.

\bibitem[\protect\citeauthoryear{Brewka, Eiter, and Truszczynski}{Brewka
  et~al\mbox{.}}{2011}]{DBLP:journals/cacm/BrewkaET11}
{\sc Brewka, G.}, {\sc Eiter, T.}, {\sc and} {\sc Truszczynski, M.} 2011.
\newblock Answer set programming at a glance.
\newblock {\em Communications of the {ACM}\/}~{\em 54,\/}~12, 92--103.

\bibitem[\protect\citeauthoryear{Calimeri, Faber, Gebser, Ianni, Kaminski,
  Krennwallner, Leone, Maratea, Ricca, and Schaub}{Calimeri
  et~al\mbox{.}}{2020}]{DBLP:journals/tplp/CalimeriFGIKKLM20}
{\sc Calimeri, F.}, {\sc Faber, W.}, {\sc Gebser, M.}, {\sc Ianni, G.}, {\sc
  Kaminski, R.}, {\sc Krennwallner, T.}, {\sc Leone, N.}, {\sc Maratea, M.},
  {\sc Ricca, F.}, {\sc and} {\sc Schaub, T.} 2020.
\newblock Asp-core-2 input language format.
\newblock {\em TPLP\/}~{\em 20,\/}~2, 294--309.

\bibitem[\protect\citeauthoryear{Calimeri, Fusc{\`{a}}, Perri, and
  Zangari}{Calimeri et~al\mbox{.}}{2017}]{DBLP:conf/aiia/CalimeriFPZ16}
{\sc Calimeri, F.}, {\sc Fusc{\`{a}}, D.}, {\sc Perri, S.}, {\sc and} {\sc
  Zangari, J.} 2017.
\newblock {I-DLV:} the new intelligent grounder of {DLV}.
\newblock {\em Intelligenza Artificiale\/}~{\em 11,\/}~1, 5--20.

\bibitem[\protect\citeauthoryear{Calimeri, Ianni, Pacenza, Perri, and
  Zangari}{Calimeri et~al\mbox{.}}{2019}]{CALIMERI_2019}
{\sc Calimeri, F.}, {\sc Ianni, G.}, {\sc Pacenza, F.}, {\sc Perri, S.}, {\sc
  and} {\sc Zangari, J.} 2019.
\newblock Incremental answer set programming with overgrounding.
\newblock {\em TPLP\/}~{\em 19,\/}~5-6 (Sep), 957–973.

\bibitem[\protect\citeauthoryear{Carbone, Katsifodimos, Ewen, Markl, Haridi,
  and Tzoumas}{Carbone et~al\mbox{.}}{2015}]{carbone2015apache}
{\sc Carbone, P.}, {\sc Katsifodimos, A.}, {\sc Ewen, S.}, {\sc Markl, V.},
  {\sc Haridi, S.}, {\sc and} {\sc Tzoumas, K.} 2015.
\newblock Apache flink{\texttrademark}: Stream and batch processing in a single
  engine.
\newblock {\em {IEEE} Data Eng. Bull.\/}~{\em 38,\/}~4, 28--38.

\bibitem[\protect\citeauthoryear{Dell'Aglio, Valle, van Harmelen, and
  Bernstein}{Dell'Aglio
  et~al\mbox{.}}{2017}]{DBLP:journals/datasci/DellAglioVHB17}
{\sc Dell'Aglio, D.}, {\sc Valle, E.~D.}, {\sc van Harmelen, F.}, {\sc and}
  {\sc Bernstein, A.} 2017.
\newblock Stream reasoning: {A} survey and outlook.
\newblock {\em Data Sci.\/}~{\em 1,\/}~1-2, 59--83.

\bibitem[\protect\citeauthoryear{Do, Loke, and Liu}{Do
  et~al\mbox{.}}{2011}]{DBLP:conf/ai/DoLL11}
{\sc Do, T.~M.}, {\sc Loke, S.~W.}, {\sc and} {\sc Liu, F.} 2011.
\newblock Answer set programming for stream reasoning.
\newblock In {\em Canadian Conference on {AI}}. LNCS, vol. 6657. Springer,
  104--109.

\bibitem[\protect\citeauthoryear{Eiter, Ogris, and Schekotihin}{Eiter
  et~al\mbox{.}}{2019}]{eiter2019distributed}
{\sc Eiter, T.}, {\sc Ogris, P.}, {\sc and} {\sc Schekotihin, K.} 2019.
\newblock A distributed approach to {LARS} stream reasoning (system paper).
\newblock {\em TPLP\/}~{\em 19,\/}~5-6, 974--989.

\bibitem[\protect\citeauthoryear{Gebser, Grote, Kaminski, and Schaub}{Gebser
  et~al\mbox{.}}{2011}]{DBLP:conf/lpnmr/GebserGKS11}
{\sc Gebser, M.}, {\sc Grote, T.}, {\sc Kaminski, R.}, {\sc and} {\sc Schaub,
  T.} 2011.
\newblock Reactive answer set programming.
\newblock In {\em {LPNMR}}. LNCS, vol. 6645. Springer, 54--66.

\bibitem[\protect\citeauthoryear{Gebser, Kaminski, Kaufmann, and Schaub}{Gebser
  et~al\mbox{.}}{2019}]{DBLP:journals/tplp/GebserKKS19}
{\sc Gebser, M.}, {\sc Kaminski, R.}, {\sc Kaufmann, B.}, {\sc and} {\sc
  Schaub, T.} 2019.
\newblock Multi-shot {ASP} solving with clingo.
\newblock {\em TPLP\/}~{\em 19,\/}~1, 27--82.

\bibitem[\protect\citeauthoryear{Gebser, Leone, Maratea, Perri, Ricca, and
  Schaub}{Gebser et~al\mbox{.}}{2018}]{DBLP:conf/ijcai/GebserLMPRS18}
{\sc Gebser, M.}, {\sc Leone, N.}, {\sc Maratea, M.}, {\sc Perri, S.}, {\sc
  Ricca, F.}, {\sc and} {\sc Schaub, T.} 2018.
\newblock Evaluation techniques and systems for answer set programming: a
  survey.
\newblock In {\em {IJCAI}}. ijcai.org, 5450--5456.

\bibitem[\protect\citeauthoryear{Hoeksema and Kotoulas}{Hoeksema and
  Kotoulas}{2011}]{hoeksema2011high}
{\sc Hoeksema, J.} {\sc and} {\sc Kotoulas, S.} 2011.
\newblock High-performance distributed stream reasoning using s4.
\newblock In {\em Ordring Workshop at ISWC}.

\bibitem[\protect\citeauthoryear{Hueske and Kalavri}{Hueske and
  Kalavri}{2019}]{hueske2019stream}
{\sc Hueske, F.} {\sc and} {\sc Kalavri, V.} 2019.
\newblock {\em Stream Processing with Apache Flink: Fundamentals,
  Implementation, and Operation of Streaming Applications}.
\newblock O'Reilly Media, Incorporated.

\bibitem[\protect\citeauthoryear{Ianni, Pacenza, and Zangari}{Ianni
  et~al\mbox{.}}{2020}]{DBLP:journals/tplp/IanniPZ20}
{\sc Ianni, G.}, {\sc Pacenza, F.}, {\sc and} {\sc Zangari, J.} 2020.
\newblock Incremental maintenance of overgrounded logic programs with tailored
  simplifications.
\newblock {\em TPLP\/}~{\em 20,\/}~5, 719--734.

\bibitem[\protect\citeauthoryear{Karimov, Rabl, Katsifodimos, Samarev,
  Heiskanen, and Markl}{Karimov
  et~al\mbox{.}}{2018}]{DBLP:conf/icde/KarimovRKSHM18}
{\sc Karimov, J.}, {\sc Rabl, T.}, {\sc Katsifodimos, A.}, {\sc Samarev, R.},
  {\sc Heiskanen, H.}, {\sc and} {\sc Markl, V.} 2018.
\newblock Benchmarking distributed stream data processing systems.
\newblock In {\em 34th {IEEE} International Conference on Data Engineering,
  {ICDE} 2018, Paris, France, April 16-19, 2018}. {IEEE} Computer Society,
  1507--1518.

\bibitem[\protect\citeauthoryear{Mileo, Abdelrahman, Policarpio, and
  Hauswirth}{Mileo et~al\mbox{.}}{2013}]{DBLP:conf/rr/MileoAPH13}
{\sc Mileo, A.}, {\sc Abdelrahman, A.}, {\sc Policarpio, S.}, {\sc and} {\sc
  Hauswirth, M.} 2013.
\newblock Streamrule: {A} nonmonotonic stream reasoning system for the semantic
  web.
\newblock In {\em {RR}}. LNCS, vol. 7994. Springer, 247--252.

\bibitem[\protect\citeauthoryear{Pham, Ali, and Mileo}{Pham
  et~al\mbox{.}}{2019}]{DBLP:conf/lpnmr/PhamAM19}
{\sc Pham, T.}, {\sc Ali, M.~I.}, {\sc and} {\sc Mileo, A.} 2019.
\newblock {C-ASP:} continuous asp-based reasoning over {RDF} streams.
\newblock In {\em {LPNMR}}. LNCS, vol. 11481. Springer, 45--50.

\bibitem[\protect\citeauthoryear{Phuoc, Dao{-}Tran, Parreira, and
  Hauswirth}{Phuoc et~al\mbox{.}}{2011}]{DBLP:conf/semweb/PhuocDPH11}
{\sc Phuoc, D.~L.}, {\sc Dao{-}Tran, M.}, {\sc Parreira, J.~X.}, {\sc and} {\sc
  Hauswirth, M.} 2011.
\newblock A native and adaptive approach for unified processing of linked
  streams and linked data.
\newblock In {\em International Semantic Web Conference {(1)}}. LNCS, vol.
  7031. Springer, 370--388.

\bibitem[\protect\citeauthoryear{Ren, Cur{\'{e}}, Naacke, and Xiao}{Ren
  et~al\mbox{.}}{2018}]{DBLP:conf/bigdataconf/RenCNX18}
{\sc Ren, X.}, {\sc Cur{\'{e}}, O.}, {\sc Naacke, H.}, {\sc and} {\sc Xiao, G.}
  2018.
\newblock Bigsr: real-time expressive {RDF} stream reasoning on modern big data
  platforms.
\newblock In {\em {IEEE} BigData}. {IEEE}, 811--820.

\bibitem[\protect\citeauthoryear{Toshniwal, Taneja, Shukla, Ramasamy, Patel,
  Kulkarni, Jackson, Gade, Fu, Donham, Bhagat, Mittal, and Ryaboy}{Toshniwal
  et~al\mbox{.}}{2014}]{DBLP:conf/sigmod/ToshniwalTSRPKJGFDBMR14}
{\sc Toshniwal, A.}, {\sc Taneja, S.}, {\sc Shukla, A.}, {\sc Ramasamy, K.},
  {\sc Patel, J.~M.}, {\sc Kulkarni, S.}, {\sc Jackson, J.}, {\sc Gade, K.},
  {\sc Fu, M.}, {\sc Donham, J.}, {\sc Bhagat, N.}, {\sc Mittal, S.}, {\sc and}
  {\sc Ryaboy, D.~V.} 2014.
\newblock Storm@twitter.
\newblock In {\em {SIGMOD} Conference}. {ACM}, 147--156.

\end{thebibliography}

%
%
%
%
%


\end{document}